\documentclass[lettersize,onecolumn]{IEEEtran}
\usepackage{amsmath,amsfonts}
\usepackage{array}
\usepackage[caption=false,font=normalsize,labelfont=sf,textfont=sf]{subfig}
\usepackage{textcomp}
\usepackage{stfloats}
\usepackage{url}
\usepackage{verbatim}
\usepackage{graphicx}
\usepackage{cite}
\usepackage[utf8]{inputenc}
\RequirePackage[OT1]{fontenc}
\RequirePackage{amsthm,amsmath}
\RequirePackage[colorlinks,citecolor=blue,urlcolor=blue]{hyperref}
\usepackage{amsmath, amssymb, amsfonts, bm, mathtools, fixmath, mathrsfs}
\usepackage{graphicx}
\usepackage{adjustbox}
\usepackage{enumerate}
\usepackage{latexsym}
\usepackage{amssymb}
\usepackage{multirow}
\usepackage{float}
\usepackage{bm}
\usepackage{paralist}
\usepackage{mathabx}
\usepackage[ruled,vlined]{algorithm2e}
\usepackage{commath}
\usepackage{comment}
\usepackage{subcaption}
\usepackage[margin=1in]{geometry}
\usepackage{subcaption}
\usepackage[inline]{enumitem}
\usepackage[figuresright]{rotating}
\usepackage{pifont}
\usepackage{threeparttable}

\numberwithin{equation}{section}

\def\bbR{\mathbb{R}}

\def \mcT{\mathcal{T}}

\def \mcS{\mathcal{S}}

\usepackage[dvipsnames]{xcolor}

\def \mcA{\mathcal{A}}

\def \bfM{\bm{M}}
\def \bfQ{\bm{Q}}
\def \bfR{\bm{R}}

\def \ev{\mathbb{E}}

\def \bfR{\mathbf{R}}

\newcommand{\vo}{\vec{o}\@ifnextchar{^}{\,}{}}

\newtheorem{innercustomgeneric}{\customgenericname}
\providecommand{\customgenericname}{}
\newcommand{\newcustomtheorem}[2]{%
  \newenvironment{#1}[1]
  {%
   \renewcommand\customgenericname{#2}%
   \renewcommand\theinnercustomgeneric{##1}%
   \innercustomgeneric
  }
  {\endinnercustomgeneric}
}
\newcustomtheorem{customthm}{Theorem}
\newcustomtheorem{customlemma}{Lemma}
\newcustomtheorem{customassumption}{Assumption}
\newtheorem{corollary}{\indent Corollary}

\newtheorem{theorem}{\indent Theorem}
\newtheorem*{theorem*}{\indent Theorem}

\theoremstyle{definition}

\NewDocumentCommand{\ceil}{s O{} m}{%
    {#2\lceil#3#2\rceil} 
}

\begin{document}

\title{Asymptotic Analysis of Sample-averaged Q-learning}

\author{Saunak Kumar Panda,  Ruiqi Liu, and Yisha Xiang
\thanks{Saunak Kumar Panda and Yisha Xiang are both with University of Houston, Houston, TX, USA.
}
\thanks{Ruiqi Liu is with Texas Tech University, Lubbock, TX, USA.}}

\markboth{IEEE TRANSACTIONS ON INFORMATION THEORY}%
{Panda \MakeLowercase{\textit{et al.}}: Asymptotic Analysis of Sample-averaged Q-learning}

\IEEEpubid{0000--0000~\copyright~2023 IEEE}

\maketitle

\begin{abstract}
Reinforcement learning (RL) has emerged as a key approach for training agents in complex and uncertain environments. Incorporating statistical inference in RL algorithms is essential for understanding and managing uncertainty in model performance. This paper introduces a generalized framework for time-varying batch-averaged Q-learning, termed sample-averaged Q-learning (SA-QL), which extends traditional single-sample Q-learning by aggregating samples of rewards and next states to better account for data variability and uncertainty. We leverage the functional central limit theorem (FCLT) to establish a novel framework that provides insights into the asymptotic normality of the sample-averaged algorithm under mild conditions. Additionally, we develop a random scaling method for interval estimation, enabling the construction of confidence intervals without requiring extra hyperparameters. Extensive numerical experiments across classic stochastic OpenAI Gym environments, including windy gridworld and slippery frozenlake, demonstrate how different batch scheduling strategies affect learning efficiency, coverage rates, and confidence interval widths. This work establishes a unified theoretical foundation for sample-averaged Q-learning, providing insights into effective batch scheduling and statistical inference for RL algorithms. 
\end{abstract}


\section{Introduction}
Reinforcement learning (RL) has emerged as a powerful tool for training agents in complex and uncertain environments \cite{sutton2018reinforcement}, \cite{arulkumaran2017deep}. However, integrating statistical inference techniques into RL is crucial for understanding the uncertainty associated with model performance and preparing RL algorithms for real-world deployment. This need arises from the stochastic nature of RL environments, where uncertainty can stem from sensor noise, unobservable states, or the agent’s interactions with the environment \cite{kober2013reinforcement}, \cite{murphy2012machine}. Statistical inference (SI) enhances decision-making under uncertainty by providing confidence intervals that define the range within which a parameter is likely to fall. This quantification of uncertainty is essential for making informed decisions and is applicable across various fields, including RL.

Confidence interval estimation is vital in RL as it offers insights into the reliability and precision of learned parameters, such as Q-values. In environments characterized by noisy data and high variability, confidence intervals help quantify the uncertainty surrounding these estimates. For example, in finance, confidence interval estimation in RL can be particularly valuable when developing trading algorithms. Consider an RL agent optimizing a stock trading strategy. The agent updates its Q-values based on historical returns and market movements, which are often noisy and volatile. Confidence intervals around these Q-values can indicate the uncertainty in expected future returns, allowing the agent to evaluate the reliability of its trading policy. Wide confidence intervals may signal high uncertainty in return estimates, prompting the agent to adopt more conservative or exploratory strategies. Conversely, narrow confidence intervals suggest greater confidence in the strategy's effectiveness, guiding the agent toward more decisive trading actions. By incorporating confidence intervals, RL applications gain enhanced insights into the reliability of learned parameters, leading to more accurate and precise model evaluations and predictions.

Traditional methods often employ the central limit theorem (CLT) to analyze algorithms under i.i.d. noise, offering valuable insights into their asymptotic behavior and performance bounds. Under fairly broad conditions, a CLT for stochastic approximation algorithms, specifically temporal-difference (TD) algorithms, characterizes their limiting variance as demonstrated by Konda \cite{konda2003}, and Devraj and Meyn \cite{devraj2018fastestconvergenceqlearning}. The CLT has been instrumental in understanding the performance of RL algorithms in i.i.d. noise environments, helping to establish convergence properties and error rates. 

Building on this foundation, a promising area of research in interval estimation has been the application of the functional central limit theorem (FCLT) to Q-learning and other RL methods. The FCLT provides a theoretical basis for the normality of certain functionals of random processes, offering a novel perspective on the statistical properties of Q-learning and similar algorithms \cite{Billingsley1999}, \cite{VanderVaart1998}. Applying FCLT results to Q-learning can enhance our understanding of its convergence behavior and uncertainty quantification.

Although there has been work on confidence interval estimation for Q-learning and other algorithms using the last iterate or final learned values \cite{Borkar2009}, \cite{Gadat2018}, \cite{borkar2024odemethodasymptoticstatistics}, asymptotic analysis for a general sample-averaged version of Q-learning has not been thoroughly explored. In many stochastic environments, sample-averaged RL algorithms can be quite beneficial. For instance, in financial trading, averaging rewards and state transitions can lead to more accurate predictions of market movements, thereby improving trading strategies. In robotics, where environments are often stochastic, averaging the rewards and transitions can result in more robust policy learning. Similarly, in game playing, this sample-averaging approach can facilitate a clearer evaluation of different strategies, ultimately leading to faster convergence to optimal policies. 

Inspired by previous works \cite{yu1994}, \cite{liu2023statisticalinferencestochasticgradient}, this paper establishes a general asymptotic result to improve statistical inference in reinforcement learning, focusing on a generalized sample-averaged Q-learning algorithm. We utilize the random-scaling method, introduced by Lee et al. \cite{lee2022fast}, to effectively handle complex asymptotic covariance matrices. This method constructs a covariance-like matrix from the trajectory of partial sums of iterates, facilitating robust statistical inference. It is particularly efficient for large datasets and resilient to variations in tuning parameters. Our contributions can be summarized as follows:

\begin{enumerate}
    \item We introduce a generalized batch average Q-learning algorithm, termed sample-averaged Q-learning, and prove that the proposed iterate achieves asymptotic normality under mild conditions. We demonstrate that the trajectory-averaged Q-value estimator \(\bar{Q}_T\) of sample-averaged Q-learning achieves asymptotic normality under mild conditions, with a convergence rate of \(O(N^{-1/2})\), retaining same asymptotic convergence
rate as standard Q-learning variants. This result is a generalization from traditional single-sample Q-learning to the sample-averaged framework. Our proof involves advanced techniques, including iterate error decomposition and partial sum decomposition of sample-averaged iterates.
    \item We develop a random scaling method for interval estimation that facilitates the construction of valid confidence intervals without relying on additional hyperparameters or the estimation of asymptotic variance. By incorporating a batch-scheduling parameter into our FCLT result, we derive different distributions with corresponding critical values based on this parameter.
\end{enumerate}

The remainder of this paper is organized as follows. Section 2 provides a review of existing statistical inference approaches in the literature. In Section 3, we introduce Q-learning within the framework of stochastic approximation (SA) and propose the sample-averaged Q-learning algorithm. Section 4 presents the main FCLT result for the proposed sample-averaged Q-learning algorithm, along with the online inference procedure using random scaling. It also extends the theoretical result to the linear function approximation (LFA) version of the proposed algorithm. Numerical experiments in stochastic versions of two classic RL environments are presented in Section 5 to quantify the performance and accuracy of the sample-averaged Q-learning algorithm under different batch-scheduling strategies. We provide several preliminary and main lemmas, proof details, as well as some additional numerical results in the appendix.


\section{Related Works}

In this section, we review existing literature on online learning algorithms, with a focus on statistical inference procedures and the development of confidence intervals. We examine various methods that address different types of data dependence and compare them to our current work. Algorithms such as stochastic gradient descent (SGD) and RL approaches such as Q-learning require accurate statistical inference methods to quantify uncertainty and ensure reliable parameter estimation.

Li et al. \cite{li2018statistical} introduce a statistical inference framework leveraging SGD's asymptotic properties and convergence results, offering a cost-effective method for parameter inference compared to bootstrapping. Su and Zhu \cite{su2018uncertainty} propose hierarchical incremental gradient descent (HIGrad), which improves parallelization capabilities over traditional SGD due to its flexible tree structure. For online decision-making applications, Chen et al. \cite{chen2020statistical} propose a method using asymptotic normality to construct confidence intervals for decision rules in SGD, while Chen et al. \cite{chen2022online} extend this to advanced decision rules, enhancing applicability in dynamic environments. Lee et al. \cite{lee2022fast} introduce an FCLT-based technique for SGD, demonstrating robustness through simulation.

Bootstrapping constructs confidence intervals by resampling data with replacement, directly estimating parameter uncertainty from the data distribution. Fang et al. \cite{fang2018online} introduce online bootstrap confidence intervals for SGD estimators, addressing parameter learning uncertainty. Kleiner et al. \cite{kleiner2014scalable} develop scalable bootstrap methods for large datasets and SGD applications, reducing traditional computational overhead. Ramprasad et al. \cite{ramprasad2022online} propose an online bootstrap inference method for policy evaluation in RL, adapting to non-i.i.d. and temporally correlated data.

Batch-means and plug-in estimators offer alternative approaches for estimating parameter uncertainty in stochastic gradient descent (SGD). Batch-means techniques reduce variance by dividing the SGD samples into batches, then averaging the results to obtain more stable estimates of uncertainty. In contrast, plug-in estimators directly substitute parameter estimates into theoretical statistical formulas, often leveraging asymptotic properties or the structure of the model. These methods are commonly used to compute key quantities such as standard errors and confidence intervals, providing valuable insight into the reliability of the estimated parameters. Chen et al. \cite{chen2020statistical_inference} propose a comprehensive framework involving batch-means and plug-in estimators for SGD, demonstrating that plug-in estimators achieve faster convergence, while batch-means require less storage and computation. For RL algorithms like Q-learning, complex covariance matrices pose challenges for plug-in methods. Zhu and Dong \cite{zhu2021constructing} adapt batch-means to construct asymptotically valid confidence regions, incorporating a process-level FCLT.

Our work differentiates itself from these approaches in several key ways. Although similar to \cite{lee2022fast} in employing FCLT, we focus on a general sample-averaged Q-learning algorithm, addressing the complexities of Markovian data dependence, which separates our work from existing SGD-focused methods. Unlike conventional bootstrapping methods which may be computationally intensive, {our approach utilizes FCLT and random scaling method, establishing a robust general theoretical framework and potentially reducing computational overhead.} Our work also builds on previous studies applying FCLT to RL algorithms, such as Li et al. \cite{li2023polyak} and Xie and Zhang \cite{xie2022statistical}, by proposing a varying step-size general version of batch-averaged Q-learning algorithm and offering a more generalized FCLT result that broadens asymptotic analysis and applicability.

\section{Problem Setup}
RL algorithms often rely on the action-value function, or Q-function, as a critical modeling element. The $\mathrm{Q}$-value for any state $s \in S$ and action $a$ under a given policy $\pi$ denoted as $Q^\pi(s, a)$, quantifies the expected cumulative reward starting from state $s$, taking action $a$, and subsequently following policy $\pi$. Mathematically, the $Q$-value is expressed as:
\begin{align*}
  Q^\pi(s, a)&=\mathrm{E}\left[r_t+\gamma r_{t+1}+\gamma^2 r_{t+2}+ \ldots \mid s_t=s, a_t=a ; \pi\right]  \nonumber
\end{align*}
where $r_t$ is the reward obtained at time step $t, s_t$ and $a_t$ are the state and action at time step $t$ respectively, and $\gamma$ is the discount factor. 
Building on this, the Bellman optimality operator, $\mcT$, is defined as:
\begin{align*}
   \mathcal{T} Q(s, a)=R(s, a) + \gamma \mathbb{E}_{S'\sim P(\cdot|s, a)}\max_{a'} Q(S', a') 
\end{align*}
where $R(s, a) \in \bbR^{\mcS \times \mcA}$ is the reward, $S' \in \mcS$ represents the next state and the expectation $\mathbb{E}[\cdot]$ is taken over the possible next states $S'$ and rewards $R$. This leads us to consider the following root-finding problem in the framework of stochastic approximation:
\begin{align}
    \ev \{Q(s,a) - \mcT Q(s,a)\} = 0.
\end{align}
The above equation holds if and only if $Q(s, a) = Q^*(s, a) \subset \bbR^{\mcS \times \mcA}$ for each state $s \in  \mcS$ and action $a \in \mcA$, where $Q^*(s,a)$ is the optimal Q-values for each $s \in  \mcS, a \in \mcA$. 

To solve for the optimal Q-values, one effective method is Q-learning, a widely used model-free, off-policy RL algorithm. The update rule for Q-learning, which is applied iteratively, is given by:
\begin{align*}
Q\left(s, a\right) &\leftarrow \left(1-\eta\right) Q\left(s, a\right) + \eta \left(R(s,a) + \gamma \max _a Q\left(s', a\right)\right)
\end{align*}
where $\eta$ represents the learning rate or step size. This iterative process allows the algorithm to converge toward the optimal Q-values, enabling the agent to learn the best policy through trial-and-error interactions with the environment.

In many applications, averaging over a sequence of i.i.d rewards and next states could potentially accelerate the convergence of RL algorithms. By averaging, we reduce the variance in our estimates, which could lead to more stable and reliable estimates of the expected rewards and future Q-values. Consider the scenario where we have a sequence of i.i.d. observations $\{(r_i,s'_i),i=1,2,\dotso\}$ given state action pair $(s, a)$. We generate $(S_t, A_t)\sim \mu$ at the $t$-step, we generate $B_t$ rewards $R_{t,1}(S_t, A_t),\ldots, R_{t, B_t}(S_t, A_t)$ and the associate next states $S'_{t,1}, \ldots, S'_{t,B_t}$. We consider dividing them into several batches. We define $\ev_\mu$ as the conditional expectation conditioned on $(S_t,A_t)$, denoted by $\ev\{\cdot|(S_t,A_t)\}$. This batching and averaging method enhances the learning process by leveraging multiple samples, which can lead to more accurate and less noisy updates. This technique can be particularly useful in scenarios where the variability of rewards and transitions can impede efficient learning.


For simplicity, we define $D=|\mcS \times \mcA| = SA$, and use matrices $\bfQ_t, \bfQ^*, \bar{\bfQ}_t \in \bbR^D$ to denote the evaluations of the functions $Q_t, Q^*, \bar{Q}_t$ respectively. To estimate $\bfQ^*$,  the sample-averaged Q-learning update is given by:
\begin{align}
\bfQ_{t+1}(S_t,A_t) 
&= \bfQ_t(S_t,A_t) - \eta_t \widehat{H}_t(\bfQ_t(S_t, A_t)), \label{eq:MBQL}
\end{align}
where,
\begin{align}
\widehat{H}_t(\bfQ_t(S_t, A_t)) &= \bfQ_t(S_t,A_t) - \widehat{\mcT}_{t+1}(\bfQ_t)(S_t,A_t) 
\intertext{and,}
\widehat{\mathcal{T}}_{t+1}(\bfQ_t)(S_t,A_t) &= \frac{1}{B_t}\sum_{i=1}^{B_t} R_{t,i}(S_t, A_t) + \frac{\gamma}{B_t}\sum_{i=1}^{B_t} \max_{a_i'}\bfQ_t(S'_{t,i}, a_i'). \nonumber
\end{align}
In particular, when $B_t=1$, the update reduces to single-sample Q-learning. Recall that the Bellman operator $\mathcal{T}$ is defined as
\begin{align*}
\mathcal{T}(\bfQ)(S_t,A_t) &= \mathbb{E}\left\{R(S_t, A_t)\right\} + \gamma \mathbb{E}_{S_t'\sim P(\cdot|S_t, A_t)}\max_{a'_t} \bfQ(S'_t, a'_t).
\end{align*}
Clearly, $\widehat{\mcT}_t$ is an unbiased estimate of $\mcT$.\\

In the following section, we explore the asymptotic behavior of the sample-averaged Q-learning iterate, presenting a novel generalized FCLT result and demonstrating its asymptotic normality. Additionally, we introduce a random scaling method for constructing confidence intervals for the Q-values of sample-averaged iterates.

\section{Asymptotic Results}\label{sec:4}
In this section, we delve into the asymptotic properties of the sample-averaged Q-learning algorithm. Specifically, we establish a general FCLT result based on the batch size. Moreover, we develop a random scaling approach for interval estimation.

\subsection{Functional Central Limit Theorem (FCLT) for Sample-averaged Q-learning}
In this subsection, we establish our main result, which is a general FCLT result for the
sample-averaged Q-learning algorithm. First, we express the expected value of $\widehat{H}_t(Q_t(S_t, A_t))$ as
\begin{align*}
H(\bfQ_t)(S_t,A_t) &= \ev_\mu\left\{ \widehat{H}_t(\bfQ_t(S_t,A_t)) \right\} = \bfQ_t - \bfR - \gamma \bfM(\bfQ_t),
\intertext{where,}
\qquad \bfR &= \ev_\mu\{R(S_t,A_t)\}, \\
\bfM(\bfQ_t) &= \ev_\mu\{\ev_{S_t' \sim P(\cdot|S_t,A_t)} \max_{a'}\bfQ_t(S_t',a')\}.
\end{align*}
We also define the matrix $\mathbf{G} = I - \gamma \textbf{P}\mathbf{\Pi}^{\pi^*}$, where:
\begin{itemize}
    \item $\mathbf{P} \in \mathbb{R}^{D \times S}$ is the transition matrix representing the probability transition kernel $P$,
    \item $\mathbf{\Pi}^{\pi^*} \in \mathbb{R}^{S \times D}$ is a projection matrix associated with a given policy $\pi$.
\end{itemize}


We begin our analysis by presenting two key assumptions necessary for our FCLT result. These assumptions are standard in the stochastic approximation literature and are not overly restrictive.
\begin{customassumption}{A1}\label{assumption:A1}
The following reward and regularity conditions are satisfied.
\begin{enumerate}
    \item (Uniformly bounded reward). The random reward $R$ is non-negative and uniformly bounded, i.e., $0 \leqslant R(s, a) \leqslant 1$ a.s. for all $(s, a) \in \mcS \times \mcA$.
    \item The matrix $-\mathbf{G}$ is Hurwitz.
\end{enumerate}
\end{customassumption}

\begin{customassumption}{A2}\label{assumption:A2}
    The following rate conditions are satisfied.
\begin{enumerate}
    \item The learning rate satisfies $\eta_t = \eta t^{-\rho}$ for some $\eta>0$ and $\rho \in (1/2,1).$
    \item The batch size $B_t$ satisfies $B_t = B \lceil t^\beta \rceil$ for some $\beta \in [0, 2\rho-1)$ and some positive integer $B$.
\end{enumerate}
\end{customassumption}

The conditions in Assumption \ref{assumption:A1} are closely related to the existing works. For example, the uniform boundedness condition for all random rewards in Assumption \ref{assumption:A1}(i) is a standard requirement in SA literature \cite{xie2022statistical}, \cite{li2023polyak}. Assumption \ref{assumption:A1} (ii) is a regularity condition required to keep the matrix $\mathbf{G}$ positive definite, which is commonly imposed in SA literature. Assumption \ref{assumption:A2} specifies some rate conditions on the step-size and batch-size of the model. Assumption \ref{assumption:A2}(i) requires that the step
size decays at a sufficiently slow rate, which is required for asymptotic normality \cite{xie2022statistical}, \cite{li2023polyak}. The learning rates satisfy $\sum_{t=1}^{\infty} \eta_t=\infty$ and $\sum_{t=1}^{\infty} \eta_t^2<\infty$, which was widely used in literature. Assumption \ref{assumption:A2}(ii) requires that the batch size should be increasing and diverging, but cannot grow too fast.


We now present the FCLT for sample-averaged Q-learning under the same conditions. We define the standardized partial sum processes associated with $\{\bfQ\}_{t \geq 0}$ as,
\begin{align*}
    \widehat{M}(r) = \frac{1}{\sqrt{\sum_{t=1}^T B_t^{-1}}} \sum_{t=1}^{\lceil r T\rceil}\left(\textbf{Q}_t-\textbf{Q}^*\right), 
\end{align*}
where, $r \in [0, 1]$ is the fraction of the data used to compute the partial-sum process and $\lceil \cdot \rceil$ returns the smallest integer larger than or equal to the input number.

\begin{theorem}\label{thm:THMONE}
Suppose that Assumptions $A 1$ and A2 are satisfied. Then it follows that
\begin{align}
         \widehat{M}(r) &\Rightarrow (1-\beta)^{1/2}\Omega^{1 / 2} \int_0^r u^{-\beta/2} d M(u).
     \intertext{\qquad Specifically for $\beta=0$, we obtain the following result}
     \widehat{M}(r) &\Rightarrow \Omega^{1/2} M(r).
\end{align}
\end{theorem}

Here, $\Rightarrow$ stands for the weak convergence in the Skorokhod space $C^{D}[0,1], M$ is a vector of $D$ independent standard Wiener processes on $[0,1]$, $\beta$ is specified in Assumption A2, and $\Omega=\mathbf{G}^{-1} \Sigma \mathbf{G}^{-1}$ where $\Sigma$ is defined in the appendix, and $\mathbf{G}$ is as defined in the beginning of the section. Note that we use \(\Rightarrow\) for convergence of stochastic processes and \(\xrightarrow{\mathbb{L}}\) for pointwise convergence of sequences of random variables. For ease of notation, we denote 
\begin{align*}
    M_\beta(r) &:= (1-\beta)^{1/2} \int_0^r u^{-\beta/2} d M(u).
    \intertext{So, Theorem \ref{thm:THMONE} can be rewritten as}
    \widehat{M}(r) &\Rightarrow \Omega^{1 / 2}  M_\beta(r)
\end{align*}

Theorem \ref{thm:THMONE} can be viewed as a generalization of Donsker’s theorem \cite{donsker1951invariance} to sample-averaged Q-learning iterates. Donsker's theorem is a fundamental result in probability theory, stating that the empirical distribution function of a sample of independent, identically distributed random variables converges in distribution to a Brownian bridge. Donsker's theorem provides the foundation for FCLTs in various contexts. The result in Theorem \ref{thm:THMONE} can be viewed as a 
generalization from traditional single-sample Q-learning to the sample-averaged
framework. It provides a general FCLT result for various values of the batch schedule parameter $\beta$. Specifically, for $\beta=0$, the distribution converges to the standard result, $\Omega^{1/2} M(r)$, where $M(r)$ is a vector of $D$ independent standard Wiener processes on $[0,1]$, aligning with previous works \cite{li2023polyak}. 

Our FCLT result for the sample-averaged Q-learning is novel; Theorem \ref{thm:THMONE} differentiates from earlier FCLT results \cite{Borkar2009}, \cite{borkar2024odemethodasymptoticstatistics} in two main aspects. Firstly, it generalizes Donsker's theorem for sample-averaged Q-learning and can be adapted for varying temporally-changing batch sizes. Secondly, it addresses the partial-sum process $\widehat{M}(r)$, explicitly formulating the asymptotic variance $\Omega$.
To prove our proposed FCLT, we use iterate error decomposition and partial sum decomposition of our sample-averaged iterates. The proof sketch of the same is provided in the next subsection.


\subsubsection{Proof Sketch}
We highlight our key contributions through this proof sketch for Theorem \ref{thm:THMONE}. The full proof of Theorem \ref{thm:THMONE} is provided in Appendix \ref{appendix:a}.

Let $\Delta_t = \textbf{Q}_t - \textbf{Q}^*$ and the noise term $\zeta_t = \widehat{H}_t(\bfQ_t(S_t,A_t)) - H(\bfQ_t)$. By the iteration formula in \ref{eq:MBQL}, we have
\begin{align*}
\Delta_t &= \Delta_{t-1} - \eta_t \left(H\left(\textbf{Q}_{t-1}\right) + \zeta_{t-1}\right)
\end{align*}
By adding and subtracting $\eta_t \mathbf{G} \Delta_{t-1}$, we obtain
\begin{align*}
\Delta_t &= \Delta_{t-1} - \eta_t \mathbf{G} \Delta_{t-1} - \eta_t \zeta_{t-1} - \eta_t \left(H\left(\textbf{Q}_{t-1}\right) - \mathbf{G} \Delta_{t-1}\right) \\
&= \left(I - \eta_t \mathbf{G}\right) \Delta_{t-1} - \eta_t \zeta_{t-1} - \eta_t \left(H\left(\textbf{Q}_{t-1}\right) - \mathbf{G} \Delta_{t-1}\right) 
\end{align*}


Expanding the error term up to $\Delta_0$ and taking the weighted average of $\Delta_t$, we simplify to obtain
\begin{align*}
\sum_{t=1}^{\lceil r T\rceil} \Delta_t &= \sum_{t=1}^{\lceil r T\rceil} 
\left(\prod_{j=1}^t\left(I-\eta_j \mathbf{G}\right)\right) \Delta_0 + \sum_{t=1}^{\lceil r T\rceil} \sum_{j=1}^t 
\left(\prod_{i=j+1}^t\left(I-\eta_i \mathbf{G}\right)\right) \eta_j \zeta_j \\
&\qquad+ \sum_{t=1}^{\lceil r T\rceil} \sum_{j=1}^t 
\left(\prod_{i=j+1}^t\left(I-\eta_i \mathbf{G}\right)\right) \eta_j
\left(H\left(\bfQ_{j-1}\right)-\mathbf{G} \Delta_{j-1}\right) \\
&= \underbrace{\sum_{t=1}^{\lceil r T\rceil} 
\left(\prod_{j=1}^t\left(I-\eta_j \mathbf{G}\right)\right) \Delta_0}_{S_1(r)} + \underbrace{\sum_{j=1}^{\lceil r T\rceil} 
\left(\sum_{t=j}^{\lceil r T\rceil} \prod_{i=j+1}^t 
\left(I-\eta_i \mathbf{G}\right)\right) \eta_j \zeta_j}_{S_2(r)} \\
&\qquad + \underbrace{\sum_{j=1}^{\lceil r T\rceil} 
    \left(\sum_{t=j}^T \prod_{i=j+1}^t 
    \left(I-\eta_i \mathbf{G}\right)\right) \eta_j \left(H\left(\bfQ_{j-1}\right) - \mathbf{G} \Delta_{j-1}\right)}_{S_3(r)}
\end{align*}
where the definitions of $S_i(r)$'s are clear from the context. We establish the overall asymptotic normality of the sum by proving the following claims:\\
\textbf{Claim 1:}\label{claim:1} Under Assumptions \ref{assumption:A1} and \ref{assumption:A2}, for $r \in[0,1]$, the following holds:
\begin{align}
    \sup _{r \in[0,1]}\left\|S_1(r)\right\| \leq \infty \text{\qquad and \qquad} \sup _{r \in[0,1]}\left\|S_3(r)\right\| \leq \infty
\end{align}
\textbf{Claim 2:} \label{claim:2}Under Assumptions \ref{assumption:A1} and \ref{assumption:A2}, the following holds:
\begin{align}
    S_2(r) \Rightarrow \Omega^{1/2} M_\beta(r).
\end{align}
While Claim 1 can be easily proved, proving Claim 2, which demonstrates the asymptotic normality of the term $S_2(r)$, presents two main challenges. Firstly, it is inherently challenging to show that $S_2(r)$ converges to asymptotic normality due to the weighted sum of martingale differences in $S_2(r)$. We achieve this in two steps, first by showing that $S_2(r)$ is asymptotically equivalent to the term $\sum_{j=1}^{\lceil r T \rceil} \mathbf{G}^{-1} \zeta_j$ and then showing that the average of the latter term is asymptotically normal. We achieve the former step by bounding 
\begin{align*}
    \sup _{r \in[0,1]}\left \| \sum_{j=1}^{\lceil r T \rceil}\left[\sum_{t=j}^{\lceil r T \rceil} \prod_{i=j+1}^t\left(I-\eta_i \mathbf{G}\right)\right] \eta_j \zeta_j-\sum_{j=1}^{\lceil r T \rceil} \mathbf{G}^{-1} \zeta_j \right \|,
\end{align*}
where we employ Burkholder's inequality and other inequalities to simplify the same. Secondly, both steps have additional challenges since we're dealing with varying sample-averaged iterates.

A direct consequence of Theorem \ref{thm:THMONE} with $r=1$ is the asymptotic distribution of the sample-averaged estimator via averaging, which is summarized as the following corollary.

\begin{corollary}\label{cor:1}
Under the uniform bounded reward assumption, it follows that
\begin{align*}
\frac{T}{\sqrt{\sum_{t=1}^T B_t^{-1}}}\left(\bar{\bfQ}_T-\bfQ^*\right) \xrightarrow{\mathbb{L}} N(0, \Omega)    
\end{align*}
\end{corollary}

Corollary 1 establishes the asymptotic normality of the estimator \(\bar{\bfQ}_T\). Noting that there are total \(N = \sum_{t=1}^T B_t\) observations used after \(T\) iterations, we can quantify the asymptotic covariance of \(\bar{\bfQ}_T\) by
\[
\operatorname{Var}\left(\sqrt{N} \bar{\bfQ}_T\right) \approx N \frac{\sum_{t=1}^T B_t^{-1}}{T^2} \Omega = M_T \Omega,
\]
where $M_T = N \frac{\sum_{t=1}^T B_t^{-1}}{T^2}$ is the variance multiplier. We can further derive the order-level bound for the estimation error as follows. Given the established asymptotic covariance expression and the total sample size \(N\), it follows that
\[
\|\overline{\textbf{Q}}_T - \textbf{Q}^*\| = O\left(\frac{\sqrt{\sum_{t=1}^T B_t^{-1}}}{T}\right) = O\left(T^{-\beta/2 - 1/2}\right) = O\left(N^{-1/2}\right).
\]

Thus, the SA-QL method retains the \(O(N^{-1/2})\) convergence rate in terms of total sample usage \(N\), consistent with the bounds established for various Q-learning algorithms (\cite{khamaru2024instance,wainwright2019variance,li2021sample,li2023qlearning}) and standard single-sample Q-learning. However, as noted in certain empirical studies \cite{liu2023statisticalinferencestochasticgradient}, the proposed method may converge more slowly in practice.\\

\begin{figure}[H]
    
    \centering
    \includegraphics[width=0.75\textwidth]{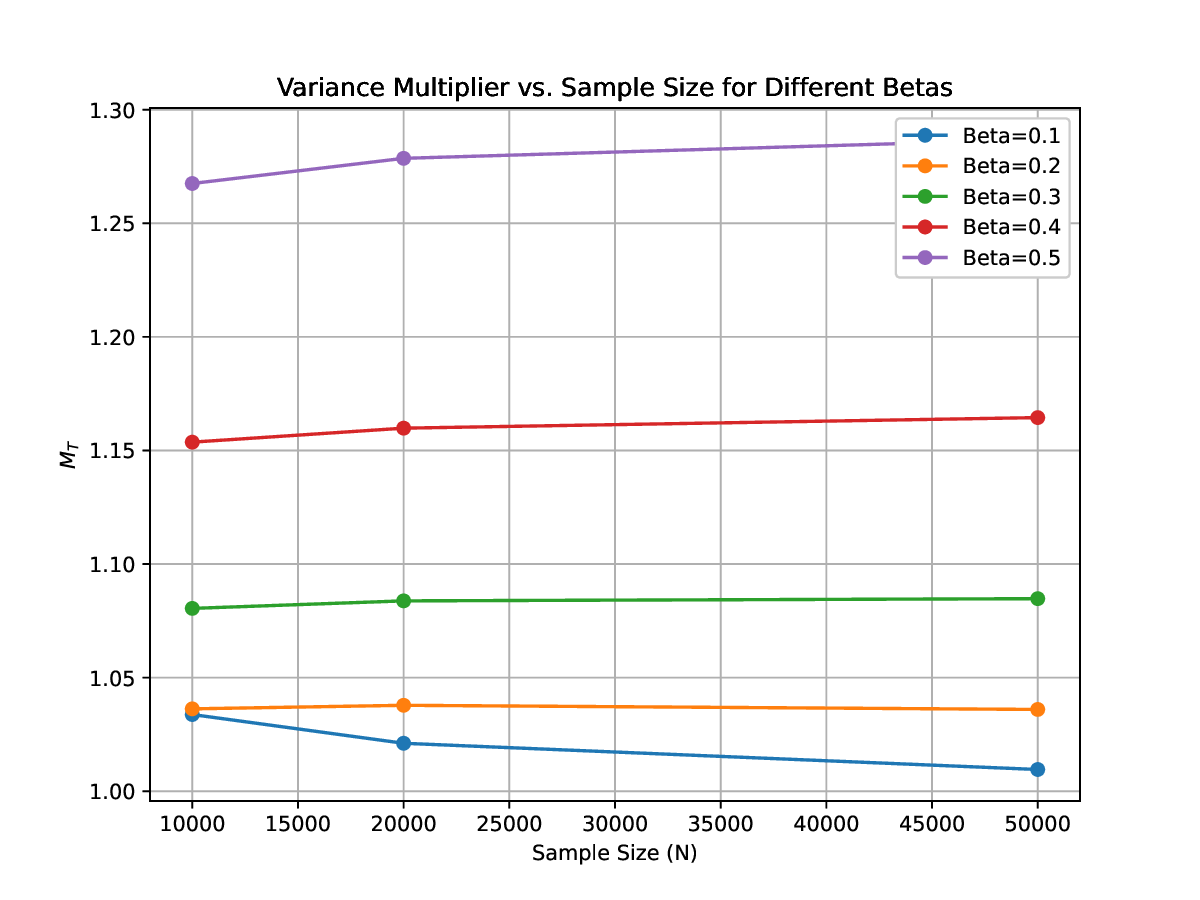}
    \caption{Variance Multiplier vs. Sample Size for Different Betas}
    \label{fig:variance_samples}
\end{figure}

To provide clearer empirical insights on the effect of the batch schedule parameter $\beta$ over variance, we analyze the variance multiplier \(M_T\) as a function of both the batch growth parameter \(\beta\) and the sample size \(N\). Figure~\ref{fig:variance_samples} illustrates how \(M_T\) varies with \(N\) for different values of \(\beta\). While increasing \(N\) results in minor fluctuations in the variance multiplier, the primary factor influencing its behavior remains \(\beta\). For smaller values of \(\beta\), \(M_T\) remains close to 1, indicating relatively stable estimation. However, as \(\beta\) increases, the variance multiplier grows, reflecting greater variability in the estimation process. This highlights the inherent tradeoff between batch size scaling and estimation accuracy, where higher batch growth reduces variance per update but increases overall variability.

The above results confirm that while SA-QL retains the same asymptotic convergence rate as standard Q-learning variants, the variance properties depend critically on $\beta$. The choice of $\beta$ influences the magnitude of confidence intervals and the efficiency of learning updates, providing the decision maker with a tunable parameter to balance learning efficiency and stability in Q-estimation.

\subsection{Online inference via Random Scaling}

In this section, we propose an online algorithm for constructing a confidence interval for $\bfQ^*$ based on Theorem \ref{thm:THMONE}. With the FCLT established for the sample-averaged algorithm, we utilize the random scaling method for inference. Originally developed for time series regression and recently adapted for stochastic gradient descent (SGD) inference \cite{lee2022fast}, the random scaling method constructs a covariance-like matrix using the entire trajectory of the partial sums of iterates. In this context, we apply the random scaling technique to studentize the partial sum of sample-averaged iterates $\widehat{M}(r)$. Before proceeding with the application, we introduce the random scaling quantity.
\begin{align*}
\widehat{D}_T &= \frac{1}{T} \sum_{s=1}^T \left\{ \frac{1}{m_T} \sum_{t=1}^s \left(\bfQ_t - \bar{\bfQ}_T\right) \right\} \left\{ \frac{1}{m_T} \sum_{t=1}^s \left(\bfQ_t - \bar{\bfQ}_T\right) \right\}^{\top}.
\end{align*}

where $m_T=\sqrt{\sum_{t=1}^T B_t^{-1}}$. 
It can be verified that $\widehat{M}(1)=T m_T^{-1}\left(\bar{\bfQ}_T-\bfQ^*\right)$ and
\begin{align*}
\widehat{D}_T=\int_0^1(\widehat{M}(r)-r \widehat{M}(1))(\widehat{M}(r)-r \widehat{M}(1))^{\top} dr .    
\end{align*}

As a consequence of the continuous mapping theorem and Theorem \ref{thm:THMONE}, we obtain the following result, which can be easily proved, and the proof is provided in appendix \ref{appendix:b}. 

\begin{theorem}\label{thm:THMTWO}
Under our bounded reward assumption, it holds that
\begin{align*}
\widehat{M}(1)^{\top} \widehat{D}_T^{-1} \widehat{M}(1)
&\xrightarrow{\mathbb{L}} M_\beta(1)^{\top} \left\{\int_0^1 \bar{M}_\beta(r) \bar{M}_\beta^{\top}(r) d r\right\}^{-1} M_\beta(1).
\end{align*}
Here $M_\beta(r)$ is d-dimensional generalized Brownian motion on $[0,1]$, and $\bar{M}_\beta(r)=M_\beta(r)-$ $r M_\beta(1)$. As a consequence, it follows that
\begin{align*}
\widehat{\kappa} = \frac{\widehat{M}(1)}{\sqrt{\widehat{D}_{T, j j}}} 
\xrightarrow{\mathbb{L}} \frac{M_{1,\beta}(1)}{\left(\int_0^1 \left\{M_{1,\beta}(r) - r M_{1,\beta}(1)\right\}^2 d r\right)^{1 / 2}} := \kappa_\beta.
\end{align*}
Here $\bar{\bfQ}_{T, j}, \bfQ_j^*$ are the $j$-th entry of $\bar{\bfQ}_T, \bfQ^*, \widehat{D}_{T, j j}$ is the $j$-th diagonal element of $\widehat{D}_T$, and $M_{1,\beta}(r)$ is a one-dimensional generalized Brownian motion.    
\end{theorem}

Theorem \ref{thm:THMTWO} shows that we can construct a statistic $\widehat{\kappa}$ that is asymptotically pivotal. Theorem \ref{thm:THMTWO} enables us to carry out statistical inference for $\bfQ^*$. As a consequence, we can construct a $(1-\alpha) \times 100 \%$-level confidence interval for $\bfQ_j^*$ as follows:
\begin{align}
\bar{\bfQ}_{T, j} \pm \kappa_{\beta,\alpha / 2} \frac{m_T}{T} \sqrt{\widehat{D}_{T, j j}} \label{eqn:cv}.
\end{align}

Here, $\kappa_{\beta, \alpha / 2}$ represents the upper $\alpha / 2$ quantile of the random variable $\kappa_\beta$. As established by Abadir and Paruolo \cite{abadir1997two}, $\kappa_\beta$ follows a mixed normal distribution that is symmetric around zero. We compute the critical values for \(\kappa_{\beta}\), as defined in Equation \ref{eqn:cv}, through simulations. Specifically, we approximate the standard Wiener process using normalized sums of i.i.d. \(N(0, 1)\) pseudo-random deviates, with 1,000 steps, 50,000 replications, and an initial seed of 1,000 for the random number generator. Table \ref{tab:kappa_combined_vals} presents the asymptotic critical values for \(\kappa_{\beta}\) for various $\beta$ values. The appendix \ref{appendix:d} provides the sampling distribution plots for all the $\beta$ values.
\begin{table*}[htbp]
    
    \centering
    \caption{Asymptotic critical values for $\kappa_{\beta}$ for various $\beta$ values}
    \label{tab:kappa_combined_vals}
    \small
    \begin{tabular}{cccccccccc}
    \hline
    \hline
         $\beta$ & 1.0\% &  2.5\% &  5.0\% &  10.0\% &  50.0\% &  90.0\% &  95.0\% &  97.5\% & 99.0\%\\ \hline
         0   & -8.581 &  -6.785 &  -5.335 &  -3.894 &  0.000 & 3.894 &  5.335 &  6.785 & 8.581\\
         0.05   & -8.550 &  -6.787 &  -5.323 &  -3.875 &  0.000 & 3.875 &  5.323 &  6.787 & 8.550\\
         0.2 & -8.393 &  -6.628 &  -5.197 &  -3.820 &  0.000 & 3.820 &  5.197 &  6.628 & 8.393\\
         0.3 & -8.149 &  -6.482 &  -5.101 &  -3.746 &  0.000 & 3.746 &  5.101 &  6.482 & 8.149\\
         0.4 & -7.931 &  -6.298 &  -4.944 &  -3.652 &  0.000 & 3.539 &  4.944 &  6.298 & 7.931\\
         0.5 & -7.593 &  -6.013 &  -4.751 &  -3.652 &  0.000 & 3.539 &  4.751 &  6.013 & 7.593\\ \hline
    \end{tabular}
\end{table*}

It is important to highlight the advantages of random scaling over traditional inference methods like plug-in and batch-means. Traditional estimators rely on the CLT and require the estimation of an asymptotic covariance matrix, which can introduce error and additional hyper-parameters, affecting efficiency and accuracy. In contrast, random scaling directly studentizes the partial sum to create a pivotal statistic. This approach eliminates the need for variance matrix estimation and avoids extra hyper-parameters, leading to greater accuracy and robustness compared to traditional methods.

\section{Numerical experiments}
    In this section, we compare the performance of sample-averaged Q-learning for different batch scheduling strategies across the windy gridworld and slippery frozenlake environment.     To provide a comprehensive evaluation, we structure the experiments into two sections: (i) \emph{Fixed timesteps comparison}, where all algorithms run for the same number of timesteps, allowing batch size schedules to dictate sample usage, and (ii) \emph{Fixed sample comparison}, where all methods use the same total number of samples. For all the experiments, we assess the performance of the different batch scheduling strategies based on three criteria: coverage rates for $\bfQ^*$, confidence interval widths, and sample efficiency.For inference, we selected a random subset consisting of 20\% of all state-action pairs for both environments to provide a more representative estimate of coverage and confidence interval width. Confidence intervals (CIs) are constructed using the random scaling (RS) method.  Hyperparameters for each experiment are specified individually, aiming for a nominal coverage rate of 95\%.\\ 

    In the windy gridworld and slippery frozenlake settings, each action has a one-third probability of deviating to a perpendicular direction instead of the intended path. To further enhance the stochastic nature of the tasks, random noise was introduced into the reward structure of each environment. Hence, immediate rewards at each step were perturbed with Gaussian noise $\mathcal{N}(0,\sigma^2)$, where $\sigma=2$.

\subsection{Fixed Timesteps Comparison}

For the first set of experiments, the total number of timesteps \(T\) was fixed across all methods. In this setting, batch size schedules determined the total number of samples processed during training. This approach allowed us to compare the efficiency of time-varying batch sizes against fixed batch sizes when the number of updates remained constant.

We conducted the experiments over 1,000 timesteps, initializing the learning rate at $\eta=0.1$ and applying a decay schedule with a factor of $\rho=0.67$. Tables~\ref{tab:gridworld_table_timesteps} and~\ref{tab:frozenlake_table_timesteps} present the results for the windy gridworld and slippery frozenlake environments under the fixed timesteps setting. The metrics include the final batch size at the end of the iterations $B_T \left(B_{\text{init}}\lceil T^\beta \rceil \right)$, nominal timesteps \(t_{\text{nom}}\) required to achieve 95\% coverage, the batch size at that point \(B_{t_{\text{nom}}}\)$\left(B_{\text{init}}\lceil t_\text{nom}^\beta \rceil \right)$, the corresponding sample usage \(N_{\text{nom}}\), and the resulting coverage probabilities and confidence interval (CI) widths. 

\begin{table}[H]
    
\centering
\caption{Sample-averaged Q-learning comparison on windy gridworld (fixed timesteps)}
\label{tab:gridworld_table_timesteps}
\begin{adjustbox}{width=\textwidth,center}
\begin{tabular}{lcccccccccccc}
\hline
\hline
 \multirow{2}{*}{$B_{\text{init}}$}  &\multirow{2}{*}{$\beta$}   &\multirow{2}{*}{\begin{tabular}[c]{@{}c@{}}$B_T$\end{tabular}} & \multirow{2}{*}{\begin{tabular}[c]{@{}c@{}}Nominal\\ Timesteps $(t_\text{nom})$\end{tabular}}  &\multirow{2}{*}{\begin{tabular}[c]{@{}c@{}}$B_{t_\text{nom}}$\end{tabular}}& \multirow{2}{*}{\begin{tabular}[c]{@{}c@{}}Nominal\\Samples$\left(N_{\text{nom}}\right)$\end{tabular}}& \multicolumn{3}{c}{Coverage Probability} & \multicolumn{3}{c}{CI Width} \\  \cline{7-12}
 &&&                    &       &                                                                            & 20\%        & 50\%        & 100\%        & 20\%    & 50\%    & 100\%    \\ \hline
 1        & 0   & 1  & 273  & 1  & 273  & 0.91        & 0.99        & 1.00         & 0.450    & 0.287    & 0.193     \\
 2        & 0   & 2  & 217  & 2  & 434  & 0.94        & 1.00        & 1.00         & 0.431    & 0.280    & 0.189     \\
 4        & 0   & 4  & 238  & 4  & 952 & 0.93        & 0.99        & 1.00         & 0.426    & 0.279    & 0.189     \\
 8       & 0   & 8 & 212  & 8 & 1696 & 0.94        & 1.00        & 1.00         & 0.439    & 0.284    & 0.192     \\
 16       & 0   & 16 & 247  & 16 & 3952 & 0.92        & 1.00        & 1.00         & 0.429    & 0.280    & 0.190     \\
 1        & 0.05& 2  & 265  & 2  & 529  & 0.91        & 0.99        & 1.00         & 0.437    & 0.284    & 0.192     \\
 1        & 0.2 & 4  & 254  & 4  & 740  & 0.92        & 0.99        & 1.00         & 0.426    & 0.276    & 0.186     \\
 1        & 0.3 & 8  & 244  & 6  & 1101 & 0.9        & 1.00        & 1.00         & 0.411    & 0.268    & 0.181     \\
 1        & 0.4 & 16 & 276  & 10 & 2012 & 0.92        & 0.99        & 1.00         & 0.437    & 0.283    & 0.191     \\ 
\hline
\end{tabular}
\end{adjustbox}
\end{table}

\begin{table}[H]
    
\centering
\caption{Sample-averaged Q-learning comparison on slippery frozenlake (fixed timesteps)}
\label{tab:frozenlake_table_timesteps}
\begin{adjustbox}{width=\textwidth,center}
\begin{tabular}{lcccccccccccc}
\hline
\hline
 \multirow{2}{*}{$B_{\text{init}}$}  &\multirow{2}{*}{$\beta$}   &\multirow{2}{*}{\begin{tabular}[c]{@{}c@{}}$B_T$\end{tabular}} & \multirow{2}{*}{\begin{tabular}[c]{@{}c@{}}Nominal\\ Timesteps $(t_\text{nom})$\end{tabular}}  &\multirow{2}{*}{\begin{tabular}[c]{@{}c@{}}$B_{t_\text{nom}}$\end{tabular}}& \multirow{2}{*}{\begin{tabular}[c]{@{}c@{}}Nominal\\Samples$\left(N_{\text{nom}}\right)$\end{tabular}}& \multicolumn{3}{c}{Coverage Probability} & \multicolumn{3}{c}{CI Width} \\  \cline{7-12}
 &&&                    &       &                                                                            & 20\%        & 50\%        & 100\%        & 20\%    & 50\%    & 100\%    \\ \hline
 1        & 0   & 1  & 205  & 1  & 411  & 0.94        & 1.00        & 1.00         & 5.905    & 3.768    & 2.527     \\
 2        & 0   & 2  & 151  & 2  & 629  & 0.99        & 1.00        & 1.00         & 5.975    & 3.798    & 2.543     \\
 4        & 0   & 4  & 124  & 5  & 1281 & 0.99        & 1.00        & 1.00         & 5.978    & 3.798    & 2.544     \\
 8       & 0   & 8 & 111  & 10 & 2441 & 0.99        & 1.00        & 1.00         & 6.008    & 3.812    & 2.552     \\
 16       & 0   & 16 & 111  & 20 & 3841 & 0.99        & 1.00        & 1.00         & 5.970    & 3.798    & 2.545     \\
 1        & 0.05& 2  & 165  & 2  & 752  & 0.98        & 1.00        & 1.00         & 5.948    & 3.779    & 2.533     \\
 1        & 0.2 & 4  & 166  & 4  & 857  & 0.98        & 1.00        & 1.00         & 5.931    & 3.747    & 2.504     \\
 1        & 0.3 & 8  & 162  & 6  & 1048 & 0.98        & 1.00        & 1.00         & 5.641    & 3.594    & 2.412     \\
 1        & 0.4 & 16 & 150  & 15 & 2076 & 0.99        & 1.00        & 1.00         & 5.976    & 3.799    & 2.542     \\ 
\hline
\end{tabular}
\end{adjustbox}
\end{table}

Across both environments, time-varying batch schedules with moderate growth parameters (\(\beta = 0.2, 0.3\)) consistently achieved the 95\% coverage threshold with fewer total samples compared to fixed batch sizes. For instance, in the gridworld environment, the configuration with \(B_{\text{init}} = 1\), \(\beta = 0.3\) reached nominal coverage after 244 timesteps, using only 1,101 samples, while the fixed batch size of \(B_{\text{init}} = 8\) required 1,696 samples to achieve the same result. A similar trend was observed in the frozenlake environment, where the same configuration achieved nominal coverage after 162 timesteps using 1,048 samples, compared to 2,441 samples for the fixed batch size of 8. This demonstrates that gradual batch size growth enhances sample efficiency without compromising convergence speed.

In terms of coverage rates, all methods ultimately reached the desired 95\% coverage with similar speed and performance across both time-varying and fixed batch schedules.

Confidence interval widths also reflected the benefits of adaptive batch scaling. For both environments, time-varying batch schedules produced narrower CIs at all stages of training compared to fixed batch methods. For example, in the gridworld environment, the final CI width for \(B_{\text{init}} = 1\), \(\beta = 0.3\) was 0.181, compared to 0.192 for the fixed batch size of 8. Similarly, in the frozenlake environment, the same configuration resulted in a final CI width of 2.412, compared to 2.552 for the fixed batch size of 8. These results indicate that adaptive batch scheduling resulted in improved stability of Q-value estimates.

However, excessive batch growth proved counterproductive. With \(\beta = 0.4\), batch sizes increased rapidly and required nearly double the sample count to reach nominal coverage compared to the more moderate \(\beta = 0.2\) and \(\beta = 0.3\) cases. Moreover, the final CI widths were consistently higher, reflecting increased variance in the estimation process. This behavior aligns with the trend noticed in Figure~\ref{fig:variance_samples} which depicted higher $M_T$ values for larger $\beta$ values.

\subsection{Fixed Sample Comparison}

In the fixed sample experiments, all methods were run with an identical total sample count \(N\), ensuring a fair comparison across batch scheduling strategies. The batch size at each timestep \(t\) follows the relation:
\[
B_t = B_{\text{init}} \lceil t^{\beta} \rceil,
\]
where \(B_{\text{init}}\) is the initial batch size and \(\beta\) controls the growth rate of batch sizes over time. Fixed batch schedules correspond to \(\beta = 0\), while \(\beta > 0\) represents time-varying schedules. The total number of samples used across all timesteps is:
\[
N = \sum_{t=1}^T B_t.
\]

We run the experiments for three sample size cases: 10000, 20000 and 50000 samples. For all the batch scheduling strategies, we begin with a learning rate of $\eta=0.1$ and decay it with a schedule factor $\rho=0.67$. Tables~\ref{tab:gridworld_table_samples} and~\ref{tab:frozenlake_table_samples} summarize the results for the windy gridworld and slippery frozenlake environments, respectively. We report the final batch size at the end of the iterations $B_T \left(B_{\text{init}}\lceil T^\beta \rceil \right)$, nominal timesteps \(t_{\text{nom}}\) required to achieve 95\% coverage, the batch size at that point \(B_{t_{\text{nom}}}\)$\left(B_{\text{init}}\lceil t_\text{nom}^\beta \rceil \right)$, the corresponding sample usage \(N_{\text{nom}}\), and the resulting coverage probabilities and confidence interval (CI) widths.


\begin{table}[H]
    
\centering
\caption{Sample-averaged Q-learning comparison on windy gridworld (fixed sample size)}
\label{tab:gridworld_table_samples}
\begin{adjustbox}{width=\textwidth,center}
\begin{tabular}{lcccccccccccc}
\hline
\hline
 \multirow{2}{*}{$B_{\text{init}}$}  &\multirow{2}{*}{$\beta$}   &\multirow{2}{*}{\begin{tabular}[c]{@{}c@{}}$B_T$\end{tabular}}& \multirow{2}{*}{$N$} & \multirow{2}{*}{\begin{tabular}[c]{@{}c@{}}Nominal\\ Timesteps $(t_\text{nom})$\end{tabular}}  &\multirow{2}{*}{\begin{tabular}[c]{@{}c@{}}$B_{t_\text{nom}}$\end{tabular}}& \multirow{2}{*}{\begin{tabular}[c]{@{}c@{}}Nominal\\Samples$\left(N_{\text{nom}}\right)$\end{tabular}}& \multicolumn{3}{c}{Coverage Probability} & \multicolumn{3}{c}{CI Width} \\  \cline{8-13}&&&                    &       &                                                                            && 20\%        & 50\%        & 100\%        & 20\%    & 50\%    & 100\%    \\ \hline
 \multirow{3}{*}{1}        &\multirow{3}{*}{0}   &1& 10000              & 411                                                                             &1 &411 & 1.00        & 1.00        & 1.00         & 0.127    & 0.071    & 0.046     \\
                      &&1& 20000              & 431                                                                             &1 &431 & 1.00        & 1.00        & 1.00         & 0.082    & 0.046    & 0.029     \\
                      &&1& 50000              &  531                                                                                 &1 &531 & 1.00            & 1.00            & 1.00             & 0.045        & 0.025        & 0.016         \\\hline
 \multirow{3}{*}{2}        &\multirow{3}{*}{0}   &2& 10000              & 315                                                                              &2 &629 & 0.99         & 1.00        & 1.00         & 0.190    & 0.109    & 0.071     \\
                      &&2& 20000              & 354                                                                             &2 &777 & 1.00        & 1.00        & 1.00         & 0.125    & 0.070    & 0.045     \\
                      &&2& 50000              & 421                                                                                  &2 &841 & 1.00            & 1.00            & 1.00              & 0.070        & 0.039        & 0.025         \\\hline
 \multirow{3}{*}{5}        &\multirow{3}{*}{0}   &5        & 10000              & 257                                                                              &5 &1281 & 0.97        & 1.00        & 1.00         & 0.317    & 0.191    & 0.126     \\
                      &&5& 20000              & 329                                                                              &5 &1411 & 0.99        & 1.00         & 1.00         & 0.215    & 0.189    & 0.081     \\
                      &&5& 50000              & 393                                                                                  &5 &1961 & 1.00            & 1.00            & 1.00             & 0.125        & 0.071        & 0.045         \\\hline
 \multirow{3}{*}{10}       &\multirow{3}{*}{0}   &10       & 10000              & 245                                                                              &10 &2441 & 0.91        & 0.99        & 1.00         & 0.434    & 0.282    & 0.191     \\
                      &&10& 20000              & 309                                                                              &10 &2511 & 0.98        & 1.00        & 1.00         & 0.318     & 0.191    & 0.126     \\
                      &&10& 50000              & 344                                                                                  &10 &3431 & 1.00            & 1.00            & 1.00             & 0.191         & 0.109        & 0.071         \\\hline
 \multirow{3}{*}{20}       &\multirow{3}{*}{0}   &20       & 10000              & 193                                                                              &20 &3841 & 0.78        & 0.98        & 1.00         & 0.508    & 0.387    & 0.276     \\
                      &&20& 20000              & 212                                                                              &20 &4221 & 0.94        & 1.00        & 1.00         & 0.438     & 0.283    & 0.191     \\
                      &&20& 50000              & 307                                                                                  &20 &6121 & 0.98            & 1.00            & 1.00             & 0.278         & 0.165        & 0.109         \\\hline
 \multirow{3}{*}{1}        &\multirow{3}{*}{0.05} &2       & 10000              & 376                                                                              &2 &752 & 1.00        & 1.00       & 1.00          & 0.188     & 0.108    & 0.070      \\
                      &&2& 20000              & 337                                                                              &2 &672 & 1.00        & 1.00        & 1.00         & 0.124    & 0.070    & 0.045     \\
                      &&2& 50000              & 407                                                                                   &2 &812 & 1.00             & 1.00            & 1.00             & 0.070        & 0.039        & 0.025         \\\hline
 \multirow{3}{*}{1}        &\multirow{3}{*}{0.2} &5       & 10000              & 284                                                                              &4 &857 & 0.97        & 1.00       & 1.00          & 0.292     & 0.174    & 0.114      \\
                      &&6& 20000              & 316                                                                              &4 &985 & 0.99        & 1.00        & 1.00         & 0.212    & 0.122    & 0.079     \\
                      &&7& 50000              & 353                                                                                   &4 &1133 & 1.00             & 1.00            & 1.00             & 0.130        & 0.074        & 0.048         \\\hline
 \multirow{3}{*}{1}        &\multirow{3}{*}{0.3} &9& 10000              & 236                                                                               &6 &1048 & 0.95        & 1.00        & 1.00         & 0.360    & 0.224    & 0.149     \\
                      &&11& 20000              & 302                                                                              &6 &1444 & 0.93         & 1.00        & 1.00          & 0.271    & 0.162     & 0.107     \\
                      &&13& 50000              & 422                                                                                  &7 &2193 & 0.99            & 1.00            & 1.00             & 0.184        & 0.106        &  0.069        \\\hline
 \multirow{3}{*}{1}        &\multirow{3}{*}{0.5} &25& 10000              & 207                                                                               &15 &2076 & 0.75         & 0.98        & 1.00         & 0.445    & 0.322    & 0.226     \\
                      &&31& 20000              & 272                                                                               &17 &3112 & 0.75        & 0.97        & 1.00         & 0.382    & 0.254    & 0.173     \\
                      &&42& 50000              &   328                                                                                &19 &4105 & 0.96            & 1.00            & 1.00             & 0.298        & 0.182        & 0.120          \\ \hline
\end{tabular}
\end{adjustbox}
\end{table}

\begin{figure}[h]
    
    \centering
    \includegraphics[width=1\textwidth]{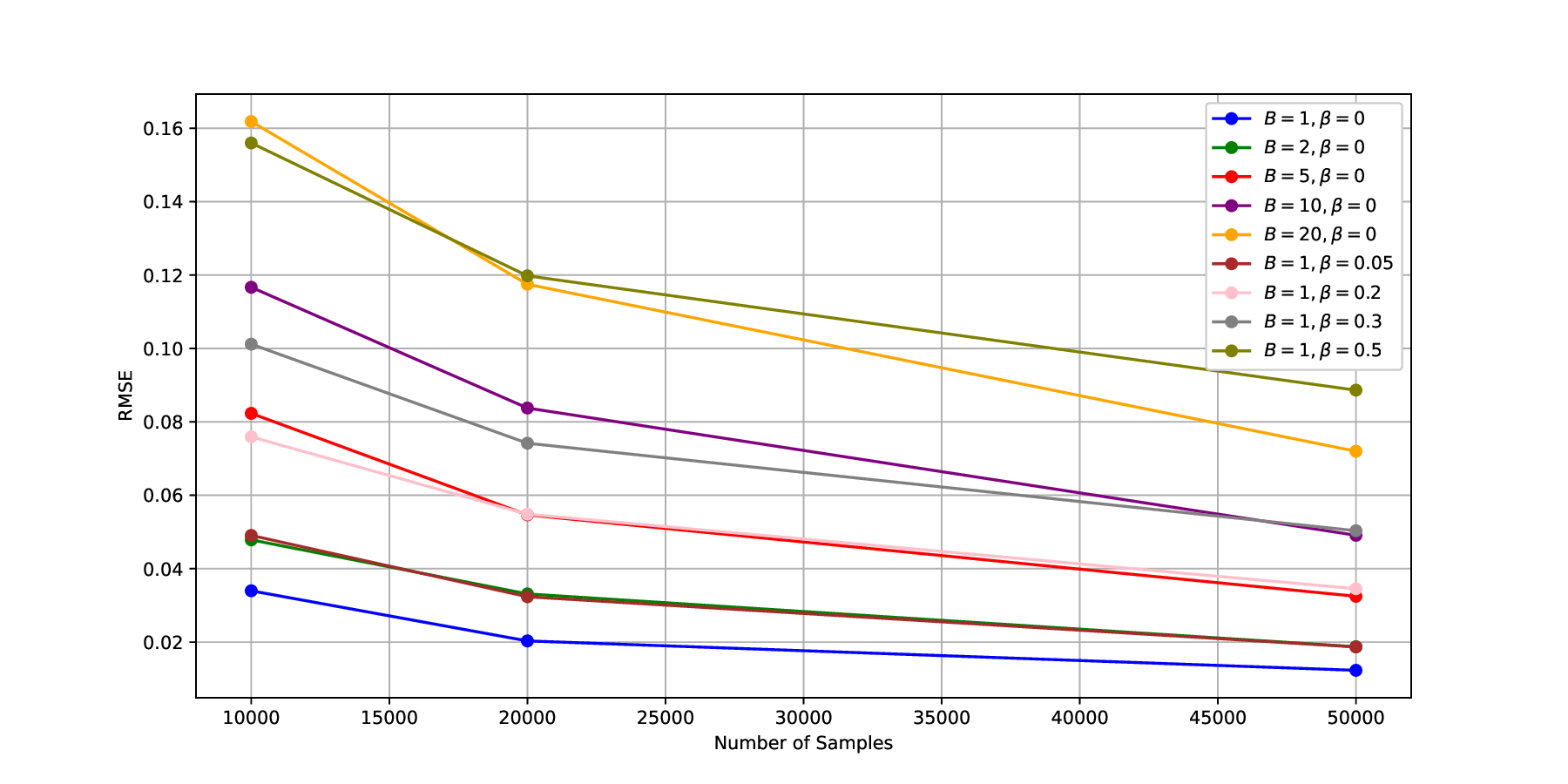}
    \caption{RMSE vs.\ Number of Samples for different $B_{\text{init}}$ and $\beta$ in the Windy Gridworld environment}
    \label{fig:rmse_gridworld}
\end{figure}

\begin{table}[H]
    
\centering
\caption{Sample-averaged Q-learning comparison on slippery frozenlake (fixed sample size)}
\label{tab:frozenlake_table_samples}
\begin{adjustbox}{width=\textwidth,center}
\begin{tabular}{lcccccccccccc}
\hline
\hline
 \multirow{2}{*}{$B_{\text{init}}$}  &\multirow{2}{*}{$\beta$}   &\multirow{2}{*}{\begin{tabular}[c]{@{}c@{}}$B_T$\end{tabular}}& \multirow{2}{*}{$N$} & \multirow{2}{*}{\begin{tabular}[c]{@{}c@{}}Nominal\\ Timesteps $(t_\text{nom})$\end{tabular}}  &\multirow{2}{*}{\begin{tabular}[c]{@{}c@{}}$B_{t_\text{nom}}$\end{tabular}}& \multirow{2}{*}{\begin{tabular}[c]{@{}c@{}}Nominal\\Samples$\left(N_{\text{nom}}\right)$\end{tabular}}& \multicolumn{3}{c}{Coverage Probability} & \multicolumn{3}{c}{CI Width} \\  \cline{8-13}&&&                    &       &                                                                            && 20\%        & 50\%        & 100\%        & 20\%    & 50\%    & 100\%    \\ \hline
 \multirow{3}{*}{1}        &\multirow{3}{*}{0}   &1& 10000              & 277                                                                             &1 &277 & 1.00        & 1.00        & 1.00         & 1.647    & 0.922    & 0.590     \\
                      &&1& 20000              & 297                                                                             &1 &297 & 1.00        & 1.00        & 1.00         & 1.069    & 0.593    & 0.378     \\
                      &&1& 50000              &  302                                                                                 &1 &302 & 1.00            & 1.00            & 1.00             & 0.591        & 0.325        & 0.206         \\\hline
 \multirow{3}{*}{2}        &\multirow{3}{*}{0}   &2& 10000              & 189                                                                              &2 &377 & 1.00         & 1.00        & 1.00         & 2.546    & 1.447    & 0.931     \\
                      &&2& 20000              & 206                                                                             &2 &411 & 1.00        & 1.00        & 1.00         & 1.658    & 0.929    & 0.594     \\
                      &&2& 50000              & 213                                                                                  &2 &425 & 1.00            & 1.00            & 1.00              & 0.930        & 0.513        & 0.327         \\\hline
 \multirow{3}{*}{5}        &\multirow{3}{*}{0}   &5        & 10000              & 129                                                                              &5 &641 & 1.00        & 1.00        & 1.00         & 4.275    & 2.532    & 1.658     \\
                      &&5& 20000              & 142                                                                              &5 &706 & 1.00        & 1.00         & 1.00         & 2.919    & 1.669    & 1.076     \\
                      &&5& 50000              & 151                                                                                  &5 &751 & 1.00            & 1.00            & 1.00             & 1.663        & 0.930        & 0.595         \\\hline
 \multirow{3}{*}{10}       &\multirow{3}{*}{0}   &10       & 10000              & 105                                                                              &10 &1041 & 0.99        & 1.00        & 1.00         & 5.981    & 3.802    & 2.545     \\
                      &&10& 20000              & 128                                                                              &10 &1271 & 1.00        & 1.00        & 1.00         & 4.293     & 2.544    & 1.663     \\
                      &&10& 50000              & 136                                                                                  &10 &1351 & 1.00            & 1.00            & 1.00             & 2.542         & 1.445        & 0.930         \\\hline
 \multirow{3}{*}{20}       &\multirow{3}{*}{0}   &20       & 10000              & 102                                                                              &20 &2021 & 0.95        & 1.00        & 1.00         & 7.267    & 5.396    & 3.792     \\
                      &&20& 20000              & 110                                                                              &20 &2181 & 0.99        & 1.00        & 1.00         & 5.989    & 3.807    & 2.548     \\
                      &&20& 50000              & 120                                                                                  &20 &2381 & 1.00            & 1.00            & 1.00             & 3.797         & 2.221        & 1.445        \\\hline
 \multirow{3}{*}{1}        &\multirow{3}{*}{0.05} &2       & 10000              & 189                                                                              &2 &376 & 1.00        & 1.00       & 1.00          & 2.547     & 1.444    & 0.927      \\
                      &&2& 20000              & 200                                                                              &2 &398 & 1.00        & 1.00        & 1.00         & 1.650    & 0.924    & 0.591     \\
                      &&2& 50000              & 201                                                                                   &2 &400 & 1.00             & 1.00            & 1.00             & 0.924        & 0.511        & 0.325         \\\hline
 \multirow{3}{*}{1}        &\multirow{3}{*}{0.2} &5       & 10000              & 183                                                                              &3 &514 & 0.99        & 1.00       & 1.00          & 3.882     & 2.289    & 1.495      \\
                      &&6& 20000              & 191                                                                              &3 &538 & 1.00        & 1.00        & 1.00         & 2.811    & 1.605    & 1.035     \\
                      &&7& 50000              & 194                                                                                   &3 &547 & 1.00             & 1.00            & 1.00             & 1.735        & 0.974        & 0.624         \\\hline
 \multirow{3}{*}{1}        &\multirow{3}{*}{0.3} &9& 10000              & 166                                                                               &5 &676 & 0.99        & 1.00        & 1.00         & 4.866    & 2.981    & 1.975     \\
                      &&11& 20000              & 193                                                                              &5 &811 & 1.00         & 1.00        & 1.00          & 3.723    & 2.184     & 1.423     \\
                      &&13& 50000              & 196                                                                                  &5 &826 & 1.00            & 1.00            & 1.00             & 2.468        & 1.405        &  0.905        \\\hline
 \multirow{3}{*}{1}        &\multirow{3}{*}{0.5} &25& 10000              & 156                                                                               &13 &1366 & 0.85         & 0.99        & 1.00         & 6.222    & 4.386    & 3.038     \\
                      &&31& 20000              & 171                                                                               &14 &1562 & 0.97        & 1.00        & 1.00         & 5.362    & 3.443    & 2.312     \\
                      &&42& 50000              &   184                                                                                &14 &1744 & 0.99            & 1.00            & 1.00             & 4.011        & 2.411        & 1.585          \\ \hline
\end{tabular}
\end{adjustbox}
\end{table}

\begin{figure}[h]
    
    \centering
    \includegraphics[width=1\textwidth]{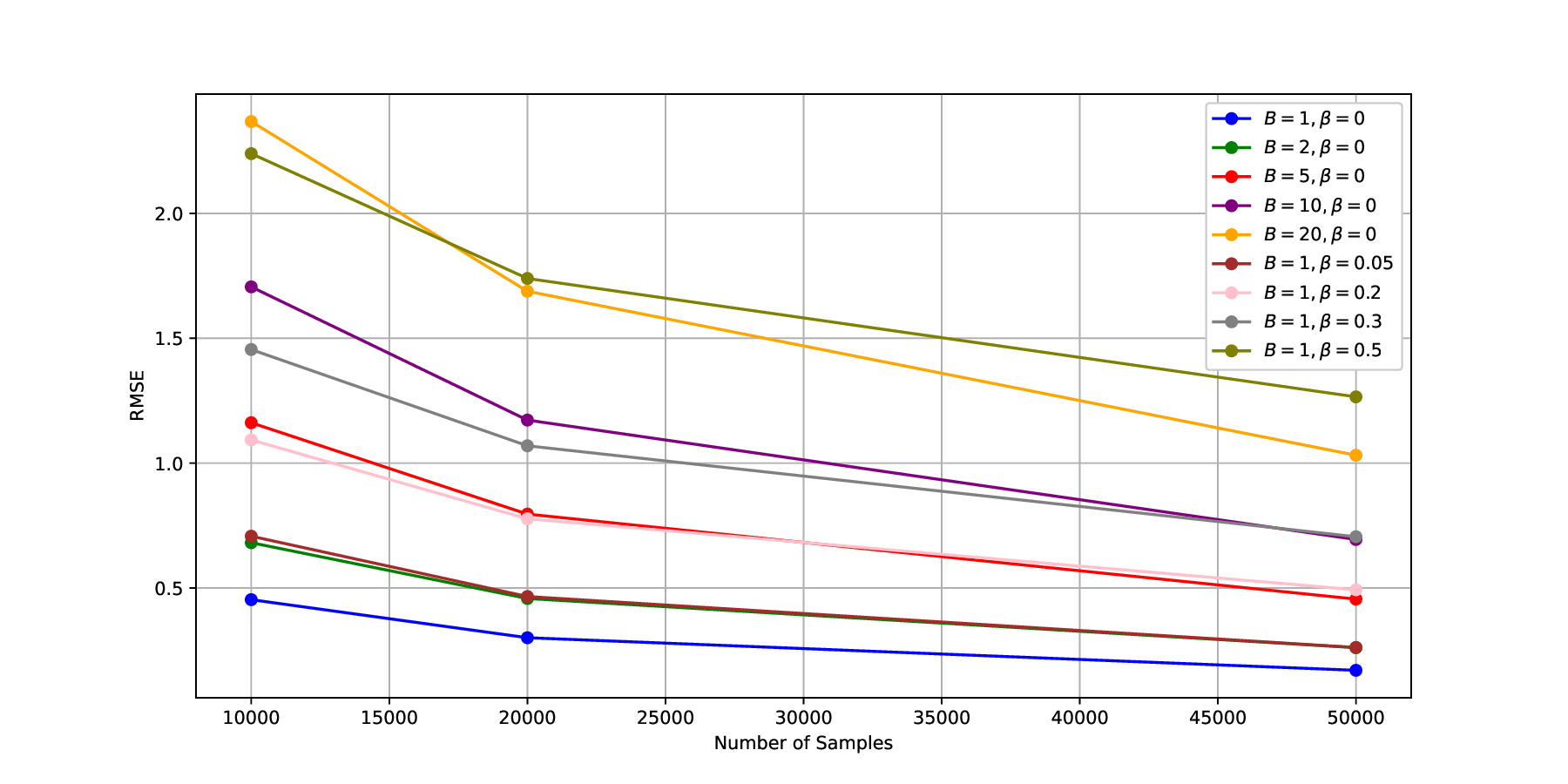}
    \caption{RMSE vs.\ Number of Samples for different $B_{\text{init}}$ and $\beta$ in the Slippery Frozenlake environment}
    \label{fig:rmse_fl}
\end{figure}



To assess sample efficiency, we compare the number of samples required to achieve nominal coverage (\(N_{\text{nom}}\)) across different batch schedules. From Tables~\ref{tab:gridworld_table_samples} and~\ref{tab:frozenlake_table_samples}, it is evident that time-varying batch schedules (\(\beta > 0\)) generally reach nominal coverage with fewer samples compared to fixed batch sizes, particularly for moderate values of \(\beta\).

For instance, in the windy gridworld environment, the configuration with \(B_{\text{init}} = 1, \beta = 0.2\) achieves nominal coverage with approximately 985 samples at \(N = 20,000\), whereas a fixed batch size of \(B_{\text{init}} = 5, \beta = 0\) requires 1411 samples to reach the same threshold. Similarly, in the frozenlake environment, the configuration \(B_{\text{init}} = 1, \beta = 0.2\) reaches coverage with 538 samples, while the fixed \(B_{\text{init}} = 5, \beta = 0\) case requires 706 samples. 


The tables further demonstrate that time-varying batch schedules (\(\beta > 0\)) achieve the 95\% coverage threshold in fewer timesteps on average compared to their fixed batch counterparts. For example, in the windy gridworld environment, the configuration \(B_{\text{init}} = 1, \beta = 0.3\) achieves nominal coverage within 236 timesteps for \(N = 10,000\), while the fixed batch \(B_t = 10\) takes 245 timesteps to reach the same threshold. 

However, overly aggressive batch growth, such as \(\beta = 0.5\), leads to slower convergence and higher sample requirements. For example, in the windy gridworld case, the configuration \(B_{\text{init}} = 1, \beta = 0.5\) requires 3,112 samples to reach coverage at \(N = 20,000\), compared to 985 samples for \(\beta = 0.2\). This confirms that while batch growth enhances sample efficiency, excessive growth can undermine early-stage learning and stability.


The confidence interval (CI) widths presented in Tables~\ref{tab:gridworld_table_samples} and~\ref{tab:frozenlake_table_samples} provide further insight into the trade-offs between batch scheduling strategies. We observed that time-varying batch schedules achieve comparable or even narrower CIs as compared to their fixed batch counterparts.

For example, in the windy gridworld environment, the configuration \(B_{\text{init}} = 1, \beta = 0.3\) achieves a CI width of 0.149 at 100\% of the training period, compared to 0.191 for the fixed batch \(B_{\text{init}} = 10, \beta = 0\). Similarly, in the frozenlake environment, the same configuration achieves a final CI width of 1.975, compared to 2.545 for the fixed batch \(B_{\text{init}} = 10, \beta = 0\).

These results suggest that moderate batch growth (\(\beta = 0.2, 0.3\)) not only improves sample efficiency and coverage rates but also achieves stable confidence intervals, outperforming fixed batch sizes in both environments. Conversely, excessive batch growth (\(\beta = 0.5\)) results in wider CIs, further showing the importance of balancing batch size growth with learning stability. These trends align with the behavior noticed in Figure~\ref{fig:variance_samples} as well.

The RMSE plots (Figures~\ref{fig:rmse_gridworld} and~\ref{fig:rmse_fl}) provide further evidence supporting the observed trade-offs between batch scheduling strategies. Across both the slippery frozenlake and windy gridworld environments, RMSE consistently decreases as the number of samples increases, reflecting improved estimation accuracy with more data.

A notable trend is that time-varying batch schedules with moderate \(\beta\) values (\(\beta = 0.2, 0.3\)) generally achieve lower RMSE compared to their fixed batch counterparts, particularly at smaller sample sizes. For example, with \(N = 10,000\), the RMSE for \(\beta = 0.2\) is lower than that for fixed batch sizes of similar final magnitude (\(B = 5, \beta = 0\)), suggesting that adaptive batch growth can facilitate more efficient early-stage learning.

As the sample size increases, RMSE for all methods converges toward similar values, though fixed batch sizes continue to exhibit slightly higher error compared to time-varying batch schedules. For higher growth rates (\(\beta = 0.5\)), RMSE reduction slows, reflecting the increased variance and inefficiency associated with overly aggressive batch growth.


\section{Conclusion}

In this paper, we introduce a generalized framework for time-varying batch-averaged
Q-learning, termed sample-averaged Q-learning. By leveraging the functional central limit theorem (FCLT), we establish a novel sample-averaged framework to account for uncertainty in data arising from different sources. We prove that the proposed algorithm achieves asymptotic normality under mild conditions and demonstrate that the trajectory-averaged Q-value estimator \(\bar{Q}_T\) of sample-averaged Q-learning achieves asymptotic normality under mild conditions, with a convergence rate of \(O(N^{-1/2})\), retaining same asymptotic convergence
rate as standard Q-learning variants. To facilitate interval estimation, we devise a random scaling (RS) method, deriving \(\kappa_\beta\) distributions with critical values tied to the batch-scheduling parameter \(\beta\). Through extensive experiments in both \emph{fixed-sample} and \emph{fixed-timesteps} settings across the slippery frozenlake and windy gridworld environments, we demonstrate that moderate batch growth strategies (\(\beta = 0.2, 0.3\)) consistently achieved nominal coverage at competitive speeds while maintaining narrower confidence intervals. The empirical results further confirmed that overly aggressive batch growth (\(\beta \geq 0.4\)) can lead to diminished sample efficiency and wider confidence intervals. This tradeoff between batch size evolution and variance reduction is effectively addressed by our asymptotic framework. Moreover, our findings reinforce the importance of appropriately tuning \(\beta\) to balance exploration, variance reduction, and sample efficiency.

We propose a tabular sample-averaged Q-learning algorithm that achieves higher inference accuracy than the single-sample version. However, tabular methods face limitations in environments with continuous or large state-action spaces due to the curse of dimensionality. While our analysis focused on synchronous Q-learning, extending the FCLT framework to asynchronous settings with Markovian noise would enhance its applicability and provide a more comprehensive evaluation. Additionally, we plan to explore hybrid approaches that combine time-varying batch sizes with explicit variance-reduction updates, aiming to improve finite-time performance while preserving the desirable asymptotic properties established in this study. A promising direction for future work is to extend the theoretical FCLT results to the Linear Function Approximation (LFA) variant of sample-averaged Q-learning and establish asymptotic normality for the LFA estimates, evaluating its performance in deep RL settings. Additionally, applying the asymptotic variance derived in this study to mean-variance risk-sensitive RL could provide insights into accuracy and performance improvements in risk-sensitive scenarios.





%

\appendix
\section{Tabular sample-averaged Q-learning proofs and additional experiments}
We present the proof of Theorem \ref{thm:THMONE}, the FCLT for the proposed sample-averaged Q-learning algorithm in this section.
\subsection{Proof of Theorem~\ref{thm:THMONE}}\label{appendix:a}
Let $\Delta_t = \textbf{Q}_t - \textbf{Q}^*$ and the noise term $\zeta_t = \widehat{H}_t(\bfQ_t(S_t,A_t)) - H(\bfQ_t)$. By the iteration formula in \ref{eq:MBQL}, we have
\begin{align*}
\Delta_t &= \Delta_{t-1} - \eta_t \left(H\left(\textbf{Q}_{t-1}\right) + \zeta_{t-1}\right)
\end{align*}
By adding and subtracting $\eta_t \mathbf{G} \Delta_{t-1}$, we obtain
\begin{align*}
\Delta_t &= \Delta_{t-1} - \eta_t \mathbf{G} \Delta_{t-1} - \eta_t \zeta_{t-1} - \eta_t \left(H\left(\textbf{Q}_{t-1}\right) - \mathbf{G} \Delta_{t-1}\right) \\
&= \left(I - \eta_t \mathbf{G}\right) \Delta_{t-1} - \eta_t \zeta_{t-1} - \eta_t \left(H\left(\textbf{Q}_{t-1}\right) - \mathbf{G} \Delta_{t-1}\right) 
\intertext{Expanding the error term till $\Delta_0$ and simplifying, we get}
         &= \left(\prod_{j=1}^t \left(I - \eta_j \mathbf{G}\right)\right) \Delta_0 + \sum_{j=1}^t \left(\prod_{i=j+1}^t \left(I - \eta_i \mathbf{G}\right)\right) \eta_j \zeta_j \\
         &\qquad  + \sum_{j=1}^t \left(\prod_{i=j+1}^t \left(I - \eta_i \mathbf{G}\right)\right) \eta_j \left(H\left(\textbf{Q}_{j-1}\right) - \mathbf{G} \Delta_{j-1}\right).
\end{align*}

Taking the weighted average of the error term $\Delta_t$ and simplifying, we show

\begin{align*}
\sum_{t=1}^{\lceil r T\rceil} \Delta_t &= \sum_{t=1}^{\lceil r T\rceil} \left(\prod_{j=1}^t \left(I - \eta_j \mathbf{G}\right)\right) \Delta_0 + \sum_{t=1}^{\lceil r T\rceil} \sum_{j=1}^t \left(\prod_{i=j+1}^t \left(I - \eta_i \mathbf{G}\right)\right) \eta_j \zeta_j \\
& \qquad +\sum_{t=1}^{\lceil r T\rceil} \sum_{j=1}^t\left(\prod_{i=j+1}^t\left(I-\eta_i \mathbf{G}\right)\right) \eta_j \left(H\left(\bfQ_{j-1}\right)-\mathbf{G} \Delta_{j-1}\right) \\
&= \underbrace{\sum_{t=1}^{\lceil r T\rceil} 
\left(\prod_{j=1}^t\left(I-\eta_j \mathbf{G}\right)\right) \Delta_0}_{S_1(r)} + \underbrace{\sum_{j=1}^{\lceil r T\rceil} 
\left(\sum_{t=j}^{\lceil r T\rceil} \prod_{i=j+1}^t 
\left(I-\eta_i \mathbf{G}\right)\right) \eta_j \zeta_j}_{S_2(r)} \\
&\qquad + \underbrace{\sum_{j=1}^{\lceil r T\rceil} 
    \left(\sum_{t=j}^T \prod_{i=j+1}^t 
    \left(I-\eta_i \mathbf{G}\right)\right) \eta_j \left(H\left(\bfQ_{j-1}\right) - \mathbf{G} \Delta_{j-1}\right)}_{S_3(r)}\\
& :=S_1(r)+S_2(r)+S_3(r),
\end{align*}
where the definitions of $S_i(r)$'s are clear from the context. We establish the overall asymptotic normality of the sum by proving the claims listed in Section \ref{sec:4}.\\

\textbf{Claim 1:}\label{claim:1} Under Assumptions \ref{assumption:A1} and \ref{assumption:A2}, for $r \in[0,1]$, the following holds:
\begin{align*}
    \sup _{r \in[0,1]}\left\|S_1(r)\right\| \leq \infty \text{\qquad and \qquad} \sup _{r \in[0,1]}\left\|S_3(r)\right\| \leq \infty
\end{align*}
\begin{proof}
    \label{appendix:a11}
We first state an auxiliary lemma useful for proving that the term $S_1(r)$ is uniformly bounded over $r \in[0,1]$. 

\begin{customlemma}{A.1}\label{lemma:A.6}
Let $\mathbf{G}$ be a positive definite matrix and define
\begin{align*}
& D_j^j=I, \\
& D_j^t=\left(I-\gamma_{t-1} \mathbf{G}\right) D_j^{t-1}=\ldots=\prod_{k-j}^{t-1}\left(I-\gamma_k \mathbf{G}\right) \quad \text{for } t \geqslant j, \\
& \bar{D}_j^t=\gamma_{j-1} \sum_{i=j}^{t-1} D_j^i=\gamma_{j-1} \sum_{i=j}^{t-1} \prod_{k-j}^{i-1}\left(I-\gamma_k \mathbf{G}\right)
\end{align*}
and the sequence $\left\{\eta_t\right\}_{t-1}^{\infty}$ satisfies $\gamma=\gamma t^{-\rho}$ for some $\gamma>0$ and $\rho \in(1 / 2,1)$. Then the following statements hold:
\begin{enumerate}
    \item There are constants $K>0$ such that $\left\|\bar{D}_j^t\right\| \leq K$ for all $j$ and all $t \geqslant j$.
    \item $\left\|D_j^t\right\| \leqslant \exp \left(-\gamma \lambda_G \sum_{k=j}^{t-1} k^{-\rho}\right)$, where $\lambda_G$ is the smallest eigenvalue of $\mathbf{G}$.
    \item If $a_j=t^b$ for some $b \in \bbR$, then it follows that $\sum_{j=1}^t a_j \mid \bar{D}_j^t-\mathbf{G}^{-1} \| \leqslant K t^{b+\rho}$ for some $K>0$.
\end{enumerate}
\end{customlemma}
\bigskip

Now, direct examination leads to
\begin{align*}
\sup _{r \in[0,1]}\left\|S_1(r)\right\| \leqslant \sum_{t=1}^T\left\|\prod_{j=1}^t\left(I-\eta_j \mathbf{G}\right)\right\| \left\|\Delta_0 \right\| 
&\stackrel{(\mathrm{i})}{\leqslant} \sum_{t=1}^T \exp \left(-c \sum_{j=1}^t j^{-\rho}\right) \\
& \asymp \sum_{t=1}^T \exp \left(-c t^{-\rho+1}\right) \\
& \asymp \int_1^T \exp \left(-c t^{-\rho+1}\right) \\
& \leqslant \int_1^{\infty} \exp \left(-c t^{-\rho+1}\right)<\infty . \qquad \tag{A.1} \label{eqn:a.1}
\end{align*}
Here (i) is due to Lemma \ref{lemma:A.6}. Similarly, we have

    \begin{align*}
\sup _{r \in[0,1]}\left\|S_3(r)\right\| & \leqslant 
\sum_{j=1}^T \left\|\sum_{t=j}^T \prod_{i=j+1}^t \left(I - \eta_i \mathbf{G}\right) \eta_j\right\| \left\|H\left(\bfQ_{j-1}\right) - \mathbf{G} \Delta_{j-1}\right\| \\
&\leqslant K \sum_{j=1}^T \left\|H\left(\bfQ_{j-1}\right) - \mathbf{G} \Delta_{j-1}\right\|
\end{align*}
where the second inequality holds due from Lemma~\ref{lemma:A.6}. 
Defining the sets:
\[
\Omega_\delta := \{j : \|\bfQ_{j-1} - \bfQ^*\| \leq \delta\}, \quad
\Omega_\delta^c := \{1, \dots, T\} \setminus \Omega_\delta.
\]
Then, decomposing the sum:
\[
\sum_{j=1}^T \left\|H(\bfQ_{j-1}) - G \Delta_{j-1} \right\| = 
\sum_{j \in \Omega_\delta} \left\|H(\bfQ_{j-1}) - G \Delta_{j-1}\right\| 
+ \sum_{j \in \Omega_\delta^c} \left\|H(\bfQ_{j-1}) - G \Delta_{j-1}\right\|.
\]
We handle these two sums separately. We now provide Lemmas~\ref{lemma:A.2} and \ref{lemma:A.11}. Lemma~\ref{lemma:A.2} asserts a local second-order bound when \(\mathbf{Q}\) is sufficiently close to \(\mathbf{Q}^\star\), and Lemma~\ref{lemma:A.11} further helps bound the second-order error term.
\begin{customlemma}{A.2}\label{lemma:A.2}
There exist constants $\delta, K > 0$ such that, for all $\mathbf{Q}$ with 
$\|\mathbf{Q} - \mathbf{Q}^\star\| \le \delta$,
\[
\bigl\|H(\mathbf{Q}) - \mathbf{G} (\mathbf{Q} - \mathbf{Q}^\star)\bigr\|
\le K \bigl\|\mathbf{Q} - \mathbf{Q}^\star\bigr\|^{2}.
\]
\end{customlemma}
\begin{customlemma}{A.3}\label{lemma:A.11}
    Under Assumptions \ref{assumption:A1} and \ref{assumption:A2}, it follows that $\mathbb{E}\left(\left\|\bfQ_t-\bfQ^*\right\|^2\right) \leqslant K \eta_t$, where $K>0$ is a constant free of $t$.
\end{customlemma}
\bigskip

Now, by Lemma~\ref{lemma:A.2}, for \(j \in \Omega_\delta\), we have,
\[
\left\|H(\bfQ_{j-1}) - G \Delta_{j-1}\right\| \leq K \|\bfQ - \bfQ^*\|^2.
\]
Now, for \(j \in \Omega_\delta^c\), we have,
\[
\left\|H(\bfQ_{j-1}) - G \Delta_{j-1}\right\| \leq \left(K' + \|G\|\right) \|\bfQ - \bfQ^*\|
\]
Now,
\[
\left\|H(\bfQ_{j-1}) - G \Delta_{j-1}\right\| \leq \max \left\{K, \frac{K' + \|G\|}{\delta} \right\}\|\bfQ - \bfQ^*\|^2, \quad \forall j
\]
Thus from above result, 
\begin{align*}
  \sum_{j=1}^T \left\|H(\bfQ_{j-1}) - G \Delta_{j-1}\right\| &\leq K \sum_{j=1}^T \left\|\Delta_{j-1}\right\|^2\\
&\stackrel{\text{(i)}}{=} O_P\left(\sum_{j=1}^T \eta_t\right) 
= O_P\left(T^{-\rho + 1}\right) 
\stackrel{\text{(ii)}}{=} o_P\left(\sqrt{\sum_{j=1}^T B_j^{-1}}\right).
\qquad \tag{A.2} \label{eqn:a.2}  
\end{align*}
Here, (i) is due to Lemma~\ref{lemma:A.11}  and (ii) is according to the rate conditions in Assumption~\ref{assumption:A2} and $2 \rho-\beta>1$.\\
Combining the bounds in (\ref{eqn:a.1}) and (\ref{eqn:a.2}), we show that $S_1(r)$ and $S_3(r)$ are uniformly bounded over $r \in[0,1]$.\\
\end{proof}

\textbf{Claim 2:} \label{claim:2}Under Assumptions \ref{assumption:A1} and \ref{assumption:A2}, the following holds:
\begin{align*}
    S_2(r) \Rightarrow \Omega^{1/2} M_\beta(r).
\end{align*}
For the $S_2(r)$ term, we first prove that it is asymptotically equivalent to the term $\sum_{j=1}^{\lceil r T \rceil} \mathbf{G}^{-1} \zeta_j$ and then show that the latter term divided by the square root of the batch sum ($\left(\sum_{j=1}^T B_j^{-1}\right))$ is asymptotically normal. Note that the matrix \(-\mathbf{G}\) plays a crucial role in the convergence analysis due to its Hurwitz property, ensuring that all its eigenvalues have negative real parts. This property guarantees that \(\mathbf{G}\) is invertible with a bounded inverse, \(\mathbf{G}^{-1}\), providing stability to the iterative process. While the norm of \(\mathbf{G}^{-1}\) does not explicitly affect the convergence analysis, its existence and boundedness are fundamental to the convergence of \(S_2(r)\).\\

\textbf{Claim 2.1:} Under Assumptions \ref{assumption:A1} and \ref{assumption:A2}, we demonstrate that
\begin{align*}
\sup _{r \in [0,1]}\left\|\sum_{j=1}^{\lceil r T \rceil} \left[\sum_{t=j}^{\lceil r T \rceil} \prod_{i=j+1}^t \left(I-\eta_i \mathbf{G}\right)\right] \eta_j \zeta_j - \sum_{j=1}^{\lceil r T \rceil} \mathbf{G}^{-1} \zeta_j \right\| = o_P\left(\sqrt{\sum_{j=1}^T B_j^{-1}}\right)
\end{align*}
\begin{proof}
Let us define matrices:
\begin{align*}
D_j^j &= I, \\
D_j^t &= \left(I - \gamma_{t-1} \mathbf{G}\right) D_j^{t-1} = \ldots = \prod_{k=j}^{t-1}\left(I - \gamma_k \mathbf{G}\right) \quad \text{for } t \geqslant j, \\
\bar{D}_j^t &= \gamma_{j-1} \sum_{i=j}^{t-1} D_j^i = \gamma_{j-1} \sum_{i=j}^{t-1} \prod_{k=j}^{i-1}\left(I - \gamma_k \mathbf{G}\right).
\end{align*}
Noting that:
\begin{align*}
\eta_j \sum_{t=j}^{\lceil r T \rceil} \prod_{i=j+1}^t\left(I - \eta_i \mathbf{G}\right)
&= \eta_j \sum_{t=j}^{\lceil r T \rceil} D_{j+1}^{t+1}
= \eta_j \sum_{i=j+1}^{\lceil r T \rceil+1} D_{j+1}^i 
= D_{j+1}^{\lceil r T \rceil+2},
\end{align*}
it follows that:
\begin{align*}
\left\|\sum_{j=1}^{\lceil r T \rceil} \left[\sum_{t=j}^{\lceil r T \rceil} \prod_{i=j+1}^t \left(I - \eta_i \mathbf{G}\right)\right] 
\eta_j \zeta_j - \sum_{j=1}^{\lceil r T \rceil} \mathbf{G}^{-1} \zeta_j \right\| 
&= \left\|\sum_{j=1}^{\lceil r T \rceil} D_{j+1}^{\lceil r T \rceil+2} \zeta_j - 
\sum_{j=1}^{\lceil r T \rceil} \mathbf{G}^{-1} \zeta_j \right\| \\
&\leqslant \left\| \sum_{j=1}^{\lceil r T \rceil} 
\left(D_{j+1}^{\lceil r T \rceil+2} - \mathbf{G}^{-1}\right) \zeta_j \right\| \\
&:= S(r).
\end{align*}
Here, the definition of \( S(r) \) is clear from the context. In the following, we will bound \( \sup_{r \in [0,1]} S(r) \). Let \( v_{s, j} = \bar{D}_{j+1}^{s+2} - \mathbf{G}^{-1} \), and we have:
\begin{align*}
\mathbb{E}\left\{\sup_{r \in [0,1]} S^{2m}(r)\right\}
= \mathbb{E}\left\{\sup_{r \in [0,1]} \left\|\sum_{j=1}^{\lceil r T \rceil} v_{\lceil r T \rceil, j} \zeta_j\right\|^{2m} \right\} 
\leqslant \sum_{s=1}^T \mathbb{E}\left\{\left\|\sum_{j=1}^s v_{s, j} \zeta_j\right\|^{2m} \right\} = \sum_{s=1}^T \mathbb{E}\left(\left\|U_s\right\|^{2m} \right), \qquad \tag{A.3} \label{eqn:a.3}
\end{align*}
where \( U_s := \sum_{j=1}^s v_{s, j} \zeta_j \) is a martingale. By direct examination, we have:
\begin{align*}
\mathbb{E}\left(\left\| U_s \right\|^{2m}\right)
= \mathbb{E}\left\{\left\|\sum_{j=1}^s v_{s, j} \zeta_j \right\|^{2m} \right\}  \stackrel{(\mathrm{i})}{\leqslant} K \mathbb{E}\left\{\left(\sum_{j=1}^s \left\|v_{s, j} \zeta_j\right\|^2 \right)^m \right\} 
= K \sum_{j_1, \ldots, j_m=1}^s \mathbb{E}\left\{\prod_{l=1}^m \left\|v_{s, j_l} \zeta_{j_l}\right\|^2 \right\}, \qquad \tag{A.4} \label{eqn:a.4}
\end{align*}
where, (i) is due to Burkholder's inequality.\\

We now provide two lemmas that establish crucial error bounds for the sample-based Bellman operator and the associated noise terms. Lemma~\ref{lemma:A.7} provides upper bounds for the approximation error of the Bellman operator under mini-batch sampling, while Lemma~\ref{lemma:A.8} further characterizes the conditional variance of the noise terms introduced by the sample-based operator. Together, these results are critical in bounding the term $S(r)$ for $r\in[0,1]$. 

\begin{customlemma}{A.4}\label{lemma:A.7}
Under Assumptions \ref{assumption:A1} and \ref{assumption:A2}, there exists a constant $K>0$ such that the following statements hold for all $\bfQ \in \Theta$ : 
\begin{enumerate}
    \item $\|H(\bfQ)(S_t,A_t)\|^2 \leqslant K$
    \item  $\mathbb{E}_\mu\left\{\left\|\widehat{H}_t(\bfQ(S_t,A_t)) - H(\bfQ_t)(S_t,A_t)\right\|^2\right\} \leqslant K B_t^{-1} $
    \item $\mathbb{E}_\mu\left\{\left\|\widehat{H}_t(\bfQ_t(S_t,A_t)) -  \widehat{H}_t(\bfQ^*(S_t,A_t)) - H(\bfQ)(S_t,A_t)\right\|^2\right\} 
    \leqslant K B_t^{-1} h_t(\bfQ)(S_t,A_t)$,\\
    where $h_t(\bfQ)(S_t,A_t):= \ev_\mu \left(\left\|\bfQ(S_t,A_t) - \mcT_{t+1} (\bfQ)(S_t,A_t) - \bfQ^*(S_t,A_t) + \mcT_{t+1} (\bfQ^*)(S_t,A_t) \right\|^2 \right)$.
\end{enumerate}
\end{customlemma}
\begin{customlemma}{A.5}\label{lemma:A.8}
    Let us define the following notations:
\begin{align*}
    \zeta_t &= \widehat{H}_t(\bfQ(S_t,A_t)) - H_t(\bfQ)(S_t,A_t) \\
    \zeta_t^* &= \widehat{H}_t(\bfQ^*(S_t,A_t) ) - H_t(\bfQ^*)(S_t,A_t) \\
    \epsilon_t &= \zeta_t - \zeta_t^*
\end{align*}
    Under Assumptions \ref{assumption:A1} and \ref{assumption:A2}, the following statements hold for some constant $K>0$.
    \begin{enumerate}
        \item $\mathbb{E}_\mu\left(\|\epsilon_t\|^2 | \mathcal{F}_{t-1}\right) \leqslant K B_t^{-1} h_{t-1}\left(\bfQ\right)$.
        \item $\mathbb{E}_\mu\left(\left\|\zeta_t^*\right\|^2\right) \leqslant K B_t^{-1}$.
        \item $\mathbb{E}_\mu \left(\| \zeta_t \|^{2 k}\right) \leqslant K B_t^{-k}$.
    \end{enumerate}
\end{customlemma}
\bigskip

Now, for each combination \( j_1, \ldots, j_m \), we see that:
\begin{align*}
\mathbb{E}\left\{\prod_{l=1}^m \left\|v_{s, j_l} \zeta_{j_l}\right\|^2 \right\}
&\leqslant \left(\prod_{l=1}^m \left\|v_{s, j_l}\right\|^2 \right) \mathbb{E}\left\{\left\| \zeta_{j_1} \right\|^2 \ldots \left\| \zeta_{j_m} \right\|^2 \right\} \\
&\stackrel{\text{(i)}}{\leqslant} K \left(\prod_{l=1}^m \left\|v_{s, j_l}\right\|^2 \right) \left(B_{j_1} \ldots B_{j_m} \right)^{-1} 
= K \prod_{l=1}^m \left(\left\|v_{s, j_l}\right\|^2 B_{j_l}\right),
\end{align*}
where \( K > 0 \) is a constant, and (i) follows from Lemma \ref{lemma:A.8}. Combining the above inequality with (\ref{eqn:a.4}), we show that:
\begin{align*}
\mathbb{E}\left(\left\|U_s\right\|^{2m}\right)
\leqslant K \sum_{j_1, \ldots, j_m=1}^s \prod_{l=1}^m \left\|v_{s, j_l}\right\|^2 B_{j_l} 
&= K \left(\sum_{j=1}^s \left\|v_{s, j}\right\|^2 B_j^{-1}\right)^m \\
& \stackrel{(\mathrm{i})}{\leqslant} K \left(\sum_{j=1}^s \left\|v_{s, j}\right\| j^{-\beta} \right)^m 
\stackrel{(\mathrm{ii})}{\leqslant} K s^{(-\beta + \rho)m}.
\end{align*}

Here (i) and (ii) are due to the boundedness of \( \|v_{s, j}\| \) and the bound of \( \sum_{j=1}^s \|v_{s, j}\| \) from Lemma \ref{lemma:A.7}. Combining the above inequality with (\ref{eqn:a.3}), we show that:
\begin{align*}
\mathbb{E}\left\{\sup _{r=[0,1]} S^{2 m}(r)\right\} 
&\leqslant K \sum_{s=1}^T s^{(-\beta+\rho) m} 
= K T^{(-\beta+\rho) m+1}.
\end{align*}

Hence, we prove that
\begin{align*}
\sup _{r=[0,1]} S(r) 
&= O_P\left(T^{\frac{1}{2 m}+\frac{\rho-\beta}{2}}\right) 
\stackrel{(\mathrm{i})}{=} o_p\left(\sqrt{\sum_{j=1}^T B_j^{-1}}\right),
\end{align*}
where (i) is due to $\sum_{j=1}^T B_j^{-1} \asymp T^{-\beta+1}$ and $m^{-1}>1-\rho$ from Assumption \ref{assumption:A2}. Hence, we complete the proof.
\end{proof}
\textbf{Claim 2.2:} Under Assumptions \ref{assumption:A1} and \ref{assumption:A2}, we show that
\begin{align*}
\left(\sum_{j=1}^T B_j^{-1}\right)^{-1 / 2} 
\sum_{j=1}^{\lfloor r T \rfloor} \mathbf{G}^{-1} \zeta_j 
&\Rightarrow \Omega^{1 / 2}  M_\beta(r)
\end{align*}
\begin{proof}
Let us define the following notations:
\begin{align*}
    \zeta_t^* &= \widehat{H}_t(\bfQ^* ) - H(\bfQ^*) \\
    \epsilon_t &= \zeta_t - \zeta_t^* \\
    m_T&=\sqrt{\sum_{j=1}^T B_j^{-1}}
\end{align*}
Let us define $X_{T, j}=m_T^{-1} \mathbf{G}^{-1} \zeta_j$. To verify the FCLT for the process $\sum_{j=1}^{\lfloor r T\rfloor} X_{T, j}$, Theorem 5.1 in \cite{merlevede2019functional} requires the following conditions:
\begin{align*}
    \max_{1 \leq j \leq T}\left|X_{T, j}\right| \text{ is uniformly integrable} \qquad \tag{1} \label{cnd:1}
    \intertext{and, for all $r \in [0,1]$,}
    \sum_{j=1}^{\lceil r T\rceil} X_{T, j} X_{T, j}^{\top} \stackrel{\mathbb{P}}{\rightarrow} r^{1-\beta}\Omega \text{ as $T \rightarrow \infty$} \qquad \tag{2} \label{cnd:2}
\end{align*}
Since $\zeta_j$ is bounded, $X_{T,j}$ is bounded and hence, Condition (\ref{cnd:1}) holds trivially.
Clearly, Condition (\ref{cnd:2}) holds trivially when $r=0$. Hence, it suffices to consider the case when $r>0$. Since,\\

\[
\zeta_t^* := \widehat{H}_t(\mathbf{Q}^*) - H(\mathbf{Q}^*),
\]
where,
\[
H(\mathbf{Q}^*) = \mathbf{Q}^* - \mathcal{T}(\mathbf{Q}^*), \quad 
\widehat{H}_t(\mathbf{Q}^*) = \mathbf{Q}^* - \widehat{\mathcal{T}}_{t+1}(\mathbf{Q}^*).
\]
Thus,
\[
\zeta_t^* = \widehat{\mathcal{T}}_{t+1}(\mathbf{Q}^*) - \mathcal{T}(\mathbf{Q}^*),
\]
where,
\[
\widehat{\mathcal{T}}_{t+1}(\mathbf{Q}^*) 
= \frac{1}{B_t} \sum_{i=1}^{B_t} \mathbf{R}_{t,i}(S_t, A_t) 
+ \frac{\gamma}{B_t} \sum_{i=1}^{B_t} \max_{a_i'} \mathbf{Q}^*(S_{t,i}', a_i'),
\]
and,
\[
\mathcal{T}(\mathbf{Q}^*) 
= \mathbf{R} + \gamma \mathbf{P}\boldsymbol{\Pi}^{\pi^\star} \mathbf{Q}^*,
\]
with,
\[
\mathbf{R} = \mathbb{E}_\mu\{\mathbf{R}(S_t, A_t)\}, \quad
\mathbf{P}\boldsymbol{\Pi}^{\pi^\star} \mathbf{Q}^*
= \mathbb{E}_\mu\{\mathbb{E}_{S_t' \sim P(\cdot \mid S_t, A_t)}[\max_{a'} \mathbf{Q}^*(S_t', a')]\}.
\]

Substituting these definitions, we get
\[
\zeta_t^* 
= \Bigl(\frac{1}{B_t} \sum_{i=1}^{B_t} \mathbf{R}_{t,i}(S_t, A_t) - \mathbf{R}\Bigr) 
+ \gamma \Bigl(\frac{1}{B_t} \sum_{i=1}^{B_t} \max_{a_i'} \mathbf{Q}^*(S_{t,i}', a_i') - \mathbf{P}\boldsymbol{\Pi}^{\pi^\star} \mathbf{Q}^*\Bigr).
\]

Now, the variance term associated with $\zeta_t^*$ is given by,
\[
\mathbb{E}_\mu\bigl[\zeta_t^*\zeta_t^{*^\top}\bigr]
= \mathbb{E}_\mu\Bigl[
  \bigl(\widehat{\mathcal{T}}_{t+1}(\mathbf{Q}^*) - \mathcal{T}(\mathbf{Q}^*)\bigr)
  \bigl(\widehat{\mathcal{T}}_{t+1}(\mathbf{Q}^*) - \mathcal{T}(\mathbf{Q}^*)\bigr)^\top
\Bigr].
\]
Substituting \(\zeta_t^*\),
\[
\mathbb{E}_\mu\bigl[\zeta_t^*\zeta_t^{*^\top}\bigr]
= \mathbb{E}_\mu\Bigl[
  \Bigl(\frac{1}{B_t} \sum_{i=1}^{B_t} \mathbf{R}_{t,i}(S_t, A_t) - \mathbf{R}\Bigr)
  \Bigl(\frac{1}{B_t} \sum_{j=1}^{B_t} \mathbf{R}_{t,j}(S_t, A_t) - \mathbf{R}\Bigr)^\top
\]
\[
+ \gamma^2 \Bigl(\frac{1}{B_t} \sum_{i=1}^{B_t} \max_{a_i'} \mathbf{Q}^*(S_{t,i}', a_i') - \mathbf{P}\boldsymbol{\Pi}^{\pi^\star} \mathbf{Q}^*\Bigr)
\Bigl(\frac{1}{B_t} \sum_{j=1}^{B_t} \max_{a_j'} \mathbf{Q}^*(S_{t,j}', a_j') - \mathbf{P}\boldsymbol{\Pi}^{\pi^\star} \mathbf{Q}^*\Bigr)^\top
\Bigr].
\]

Since the sampled rewards \(\mathbf{R}_{t,i}(S_t, A_t)\) and the \(\max \mathbf{Q}^*\) terms are independent across \(i\). Then:
\[
\frac{1}{B_t} \sum_{i=1}^{B_t} \mathbf{R}_{t,i}(S_t, A_t)
\quad \text{and} \quad 
\frac{1}{B_t} \sum_{i=1}^{B_t} \max_{a_i'} \mathbf{Q}^*(S_{t,i}', a_i')
\]
are unbiased estimates of \(\mathbf{R}\) and \(\mathbf{P}\boldsymbol{\Pi}^{\pi^\star} \mathbf{Q}^*\), respectively.\\

For the first variance term, we have
\[
\mathbb{E}_\mu\Bigl[\Bigl(\frac{1}{B_t} \sum_{i=1}^{B_t} \mathbf{R}_{t,i}(S_t, A_t) - \mathbf{R}\Bigr)
\Bigl(\frac{1}{B_t} \sum_{j=1}^{B_t} \mathbf{R}_{t,j}(S_t, A_t) - \mathbf{R}\Bigr)^\top\Bigr]
= \frac{1}{B_t}\Bigl(\mathbb{E}_\mu[\mathbf{R}\mathbf{R}^\top] - \mathbb{E}_\mu[\mathbf{R}]\mathbb{E}_\mu[\mathbf{R}]^\top\Bigr).
\]

For the second term, we have
\begin{align*}
    \mathbb{E}_\mu\Bigl[\Bigl(\frac{1}{B_t} \sum_{i=1}^{B_t} \max_{a_i'} \mathbf{Q}^*(S_{t,i}', a_i') - \mathbf{P}\boldsymbol{\Pi}^{\pi^\star} \mathbf{Q}^*\Bigr)&
\Bigl(\frac{1}{B_t} \sum_{j=1}^{B_t} \max_{a_j'} \mathbf{Q}^*(S_{t,j}', a_j') - \mathbf{P}\boldsymbol{\Pi}^{\pi^\star} \mathbf{Q}^*\Bigr)^\top\Bigr]\\
&= \frac{\gamma^2}{B_t}\Bigl(\mathbb{E}_\mu[\mathbf{M}\mathbf{M}^\top] - \mathbb{E}_\mu[\mathbf{M}]\mathbb{E}_\mu[\mathbf{M}]^\top\Bigr),
\end{align*}

where, $\mathbf{M} = \mathbf{P}\boldsymbol{\Pi}^{\pi^\star} \mathbf{Q}^*$. Combining the two terms, we get
\[
\mathbb{E}_\mu\bigl[\zeta_t^*\zeta_t^{*^\top}\bigr]
= \frac{1}{B_t} \Sigma, \qquad \tag{A.5} \label{eqn:a.5}
\]

where,
\[
\Sigma 
= \Bigl(\mathbb{E}_\mu[\mathbf{R}\mathbf{R}^\top] - \mathbb{E}_\mu[\mathbf{R}]\mathbb{E}_\mu[\mathbf{R}]^\top\Bigr)
+ \gamma^2 \Bigl(\mathbb{E}_\mu[\mathbf{M}\mathbf{M}^\top] - \mathbb{E}_\mu[\mathbf{M}]\mathbb{E}_\mu[\mathbf{M}]^\top\Bigr).
\]
Hence, by direct examination, we have
\begin{align*}
D(r) :=m_T^{-2} \sum_{j=1}^{\mid r T\rceil} \mathbf{G}^{-1} \mathbb{E}_\mu\left(\zeta_t^* \zeta_t^{*^\top}\right) \mathbf{G}^{-1} 
\stackrel{(\mathrm{i})}{=} m_T^{-2} \sum_{j=1}^{\mid r T\rceil} \mathbf{G}^{-1} \left( \frac{1}{B_t} \Sigma \right) \mathbf{G}^{-1} 
\stackrel{\text { (ii) }}{\rightarrow} r^{1-\beta} \Omega. \qquad \tag{A.6} \label{eqn:a.6}
\end{align*}
Here (i) is due to (\ref{eqn:a.5}), and (ii) follows from Lemma \ref{lemma:A.4} stated below that characterizes the limiting ratio of partial sums of inverse batch sizes for batch size following growth pattern given in Assumption~\ref{assumption:A2}.
\begin{customlemma}{A.6}\label{lemma:A.4}
Suppose that $B_t=c\ceil{t^{\beta}}$ for some $\beta\in [0, 1)$ and constant $c$, then it holds that
\begin{align*}
\lim_{T\to \infty}\frac{\sum_{t=1}^{\ceil{rT}}B_t^{-1}}{\sum_{t=1}^{T}B_t^{-1}}=r^{1-\beta}.
\end{align*} 
\end{customlemma}
\bigskip

Now, let $Y_{T, j}=X_{T, j} X_{T, j}^{\top}-\mathbb{E}_\mu\left(X_{T, j} X_{T, j}^{\top} \mid \mathcal{F}_{j-1}\right)$, and it follows that
\begin{align*}
   \mathbb{E}_\mu\left(\left\|Y_{T, j}\right\|^2\right) \leqslant 2 \mathbb{E}_\mu\left(\left\|X_{T, j}\right\|^4\right) 
   \leqslant 2 m_T^{-4} \mathbb{E}_\mu\left(\left\|\zeta_j\right\|^4\right) \stackrel{(\mathrm{i})}{\leqslant} K m_T^{-4} B_j^{-2}, 
\end{align*}
where (i) is due to Lemma \ref{lemma:A.8}. If $i<j$, we see that
\begin{align*}
    \mathbb{E}_\mu\left\{\operatorname{trace}\left(Y_{T, j}^{\top} Y_{T, i}\right) \mid \mathcal{F}_i\right\}
    =\operatorname{trace}\left\{\mathbb{E}_\mu\left(Y_{T, j}^{\top} Y_{T, i} \mid \mathcal{F}_i\right)\right\} 
    =\operatorname{trace}\left\{\mathbb{E}_\mu\left(Y_{T, j}^{\top} \mid \mathcal{F}_i\right) Y_{T, i}\right\}
    =0.
\end{align*}
Therefore, the above two inequalities together imply that
\begin{align*}
\mathbb{E}_\mu\left\{\left\|\sum_{j=1}^T Y_{T, j}\right\|^2\right\} 
=\sum_{j=1}^T \mathbb{E}_\mu\left(\left\|Y_{T, j}\right\|^2\right) + 2\sum_{1 \leqslant i<j \leqslant T} \mathbb{E}_\mu\left\{\operatorname{trace}\left(Y_{T, j}^{\top} Y_{T, i}\right) \right\} 
\leqslant K m_T^{-4} \sum_{j=1}^T B_j^{-2} \rightarrow 0,
\end{align*}
which further leads to
\begin{align*}
    \sum_{j=1}^T X_{T, j} X_{T, j}^{\top}-\sum_{j=1}^T \mathbb{E}_\mu\left(X_{T, j} X_{T, j}^{\top} \mid \mathcal{F}_{j-1}\right)&=\sum_{j=1}^T Y_{T, j} 
    =o_P(1).  \tag{A.7} \label{eqn:a.7}
\end{align*}
Moreover, direct examination shows that
\begin{align*}
\left\|\mathbb{E}\left(\zeta_j \zeta_j^{\top} \mid \mathcal{F}_{j-1}\right)-\mathbb{E}\left(\eta_j \eta_j^{\top}\right)\right\| 
&=\left\|\mathbb{E}\left(\zeta_j \zeta_j^{\top} \mid \mathcal{F}_{j-1}\right)-\mathbb{E}\left(\eta_j \eta_j^{\top} \mid \mathcal{F}_{j-1}\right)\right\| \\
& =\left\|\mathbb{E}\left(\zeta_j \zeta_j^{\top}-\eta_j \eta_j^{\top} \mid \mathcal{F}_{j-1}\right)\right\| \\
& =\left\|\mathbb{E}\left\{\left(\zeta_j-\eta_j\right)\left(\zeta_j-\eta_j\right)^{\top} +2\left(\zeta_j-\eta_j\right) \eta_j^{\top} \mid \mathcal{F}_{j-1}\right\}\right\| \\
& =\left\|\mathbb{E}\left\{\epsilon_j \epsilon_j^{\top}+2 \epsilon_j \eta_j^{\top} \mid \mathcal{F}_{j-1}\right\}\right\|. \qquad \tag{A.8} \label{eqn:a.8}
\end{align*}
To facilitate the final bound in the proof below, we rely on two auxiliary results that characterize the asymptotic behavior of the error terms arising from sample-averaged Q-learning. Lemma~\ref{lemma:A.5} ensures the convergence of weighted averages when the weights diminish appropriately, while Lemma~\ref{lemma:A.10} establishes the almost sure convergence of the Q-function residual \(h_T(\bfQ)\) to zero. Together, these results ensure that the cumulative error term vanishes asymptotically
\begin{customlemma}{A.7}\label{lemma:A.5}
    Suppose that $a_n>0$ is a decreasing sequence with $\lim _{n \rightarrow \infty} a_n=0$, and $b_n$ is $a$ converging sequence with $\lim_{n \rightarrow{} \infty} b_n=b$. If $\sum_{n=1}^{\infty} a_n=\infty$, then $\lim_{T \rightarrow \infty} \sum_{n=1}^T a_n b_n /\left(\sum_{n=1}^T a_n\right)=b$.
\end{customlemma}
\begin{customlemma}{A.8}\label{lemma:A.10}
    Let $h_T\left(\bfQ\right)$ be the error term defined in Lemma~\ref{lemma:A.7}:
    \begin{align*}
        h_t(\bfQ)(S_t,A_t):= \ev_\mu \left(\left\|\bfQ(S_t,A_t) - \mcT_{t+1} (\bfQ)(S_t,A_t) - \bfQ^*(S_t,A_t) + \mcT_{t+1} (\bfQ^*)(S_t,A_t) \right\|^2 \right).
    \end{align*}
    Under Assumptions $\ref{assumption:A1}$ and \ref{assumption:A2}, it follows that $h_T\left(\bfQ\right) \stackrel{\text { a.s. }}{\longrightarrow} 0$ as $T \rightarrow \infty$.
\end{customlemma}
\bigskip

Now, using (\ref{eqn:a.8}), we have
\begin{align*}
\left\|\sum_{j=1}^T \mathbb{E}\left(X_{T, j} X_{T, j}^{\top} \mid \mathcal{F}_{j-1}\right)-D_1(r)\right\|
& =\left\|\sum_{j=1}^T \mathbb{E}\left(X_{T, j} X_{T, j}^{\top} \mid \mathcal{F}_{j-1}\right) -\sum_{j=1}^T m_T^{-2} \mathbf{G}^{-1} \mathbb{E}\left(\eta_j \eta_j^{\top}\right) \mathbf{G}^{-1}\right\| \\
& =\left\|\sum_{j=1}^T m_T^{-2} \mathbf{G}^{-1} \mathbb{E}\left(\zeta_j \zeta_j^{\top} -\eta_j \eta_j^{\top} \mid \mathcal{F}_{j-1}\right) \mathbf{G}^{-1}\right\| \\
& \leqslant m_T^{-2}\left\|\mathbf{G}^{-1}\right\|^2 \sum_{j=1}^T\left(\mathbb{E}\left(\left\|\epsilon_j\right\|^2 | \mathcal{F}_{j-1}\right) +2 \sqrt{\mathbb{E}\left(\left\|\epsilon_j\right\|^2 | \mathcal{F}_{j-1}\right)} \sqrt{\mathbb{E}\left(\left\|\eta_j\right\|^2\right)}\right) \\
& \stackrel{(\mathrm{i})}{\leqslant} m_T^{-2} \sum_{j=1}^T\left(B_j^{-1} h_{j-1}\left(\bfQ\right)+2 B_j^{-1} \sqrt{h_{j-1}\left(\bfQ\right)}\right) \\
& =K m_T^{-2} \sum_{j=1}^T B_j^{-1}\left(h_{j-1}\left(\bfQ\right)+\sqrt{h_{j-1}\left(\bfQ\right)}\right) \\
&\stackrel{\text { (ii) }}{=} o_P(1).
\end{align*}
Here (i) is due to Lemma \ref{lemma:A.8}, and (ii) follows from Lemma \ref{lemma:A.5} and the fact $h_t\left(\bfQ\right) \stackrel{\text { a.s. }}{\longrightarrow} 0$ in Lemma \ref{lemma:A.10}. Combining the above inequality with (\ref{eqn:a.6}) and (\ref{eqn:a.7}), we verify Condition (\ref{cnd:2}). 
\end{proof}

\subsection{Proof of Theorem \ref{thm:THMTWO}}\label{appendix:b}

\begin{proof}
Recall that $m_T=\sqrt{\sum_{t=1}^T B_t^{-1}}$ and
\[
\widehat{M}(r) = m_T^{-1} \sum_{t=1}^{\lceil r T\rceil}\left(\mathbf{Q}_t - \mathbf{Q}^*\right).
\]

Noting that $\bar{\bfQ}_T-\textbf{Q}^*=m_T T^{-1} \widehat{M}(1)$ and
\begin{align*}
m_T\left\{\widehat{M}\left(\frac{s}{T}\right)-\frac{s}{T} \widehat{M}(1)\right\} & =\sum_{t=1}^s\left(\bfQ_t-\bfQ^*\right)-\frac{s}{T} \sum_{t=1}^T\left(\bfQ_t-\bfQ^*\right) \\
& =\sum_{t=1}^s\left(\bfQ_t-\bfQ^*\right)-s\left(\bar{\bfQ}_T-\bfQ^*\right)=\sum_{t=1}^s\left(\bfQ_t-\bar{\bfQ}_T\right)
\end{align*}
we have
\begin{align*}
\widehat{D}_T & =\frac{1}{T} \sum_{s=1}^T\left\{\frac{1}{m_T} \sum_{t=1}^s\left(\bfQ_t-\bar{\bfQ}_T\right)\right\}\left\{\frac{1}{m_T} \sum_{t=1}^s\left(\bfQ_t-\bar{\bfQ}_T\right)\right\}^{\top} \\
& =\frac{1}{T} \sum_{s=1}^T\left\{\widehat{M}\left(\frac{s}{T}\right)-\frac{s}{T} \widehat{M}(1)\right\}\left\{\widehat{M}\left(\frac{s}{T}\right)-\frac{s}{T} \widehat{M}(1)\right\}^{\top} \\
& =\int_0^1\left(\widehat{M}(r)-r \widehat{M}(1)\right)\left(\widehat{M}(r)-r \widehat{M}(1)\right)^{\top} d r
\end{align*}

Let us define the functionals for the process $M(\cdot)$:
\begin{align*}
\Phi_1({M}(\cdot))&= \int_0^1 \bigl[{M}(r) - r {M}(1)\bigr] \bigl[{M}(r) - r {M}(1)\bigr]^\top dr,\\
\Phi_2({M}(\cdot))&= M(1).
\end{align*}
Moreover, let $\Phi(M(\cdot))=(\Phi_1({M}(\cdot)), \Phi_2({M}(\cdot)))$. Then, we have
\begin{align*}
    \Phi(\widehat{M}(\cdot))&=(\Phi_1(\widehat{M}(\cdot)), \Phi_2(\widehat{M}(\cdot)))=(\widehat{D}_T, \widehat{M}(1)),\\
     \Phi(\Omega^{1/2}{M}_\beta(\cdot))&=( \Phi_1(\Omega^{1/2}{M}_\beta(\cdot)),  \Phi_2(\Omega^{1/2}{M}_\beta(\cdot)))=\left(\int_0^1 \Omega^{1/2}\bar{M}_\beta(r) \bar{M}_\beta^{\top}(r) \Omega^{1/2}d r, \Omega^{1/2}M_\beta(1)\right).
\end{align*}
Here,
\begin{align*}
\widehat{D}_T&=\int_0^1 \bigl[\widehat{M}(r) - r \widehat{M}(1)\bigr] \bigl[\widehat{M}(r) - r \widehat{M}(1)\bigr]^\top dr,\\
    \bar{M}_\beta(r)&=M_\beta(r)- r M_\beta(1). 
\end{align*}
Since Theorem 1 establishes weak convergence of $\widehat{M}(r)$ to $\Omega^{1 / 2}M_\beta(r)$, the Continuous Mapping Theorem ensures that
\[
\Phi(\widehat{M}(\cdot)) \xrightarrow{\mathbb{L}} \Phi(\Omega^{1 / 2}M_\beta(\cdot)),
\]
which translates to
\[
(\widehat{D}_T, \widehat{M}(1)) \xrightarrow{\mathbb{L}} \left(\int_0^1 \Omega^{1/2}\bar{M}_\beta(r) \bar{M}_\beta^{\top}(r) \Omega^{1/2}d r, \Omega^{1/2}M_\beta(1)\right)\; \textrm{jointly}.
\]
Using the above joint convergence and continuous mapping theorem, we obtain that
\begin{align*}
    \widehat{M}^\top(1) \widehat{D}_T^{-1}\widehat{M}(1) &\xrightarrow{\mathbb{L}} M_\beta^\top(1)\Omega^{1/2}\left(\int_0^1 \Omega^{1/2}\bar{M}_\beta(r) \bar{M}_\beta^{\top}(r) \Omega^{1/2}d r\right)^{-1}\Omega^{1/2}M_\beta(1)\\
    &=M_\beta^\top(1)\left(\int_0^1 \bar{M}_\beta(r) \bar{M}_\beta^{\top}(r)d r\right)^{-1}M_\beta(1),
\end{align*}
which completes the proof.
\end{proof}

\subsection{Proof of Main and Preliminary Lemmas}
In this section, we provide the proofs of all the main lemmas used in the theorems and other preliminary lemmas.\\

\subsubsection{Proof of Lemma~\ref{lemma:A.6}}
\begin{proof}
    This is Lemma S.10 in \cite{liu2023statisticalinferencestochasticgradient}.
\end{proof}

\subsubsection{Proof of Lemma~\ref{lemma:A.2}}
\begin{proof}
The result is proven in \cite{li2023polyak} for $l_\infty$ norm taking $(\delta,K)=(\infty,\frac{L}{\Delta})$ provided the optimal policy $\pi^{\star}$ is unique, where $\Delta$ is the optimality gap defined as
\[
\Delta := \min_{s} \min_{a \neq \pi^{\star}(s)} \bigl| V^{\star}(s) - Q^{\star}(s, a) \bigr|.
\]
\end{proof}

\subsubsection{Proof of Lemma~\ref{lemma:A.11}}
\begin{proof}
We first state an auxiliary lemma useful in proving our result.
\begin{customlemma}{A.3.1}\label{lemma:A.9}
     Under Assumptions \ref{assumption:A1} and \ref{assumption:A2}, it follows that $\bfQ_T \stackrel{\text { a.s. }}{\longrightarrow} \bfQ^*$ as $T \rightarrow \infty$.
\end{customlemma}
\bigskip
Using (\ref{eqn:a.9}) from proof of Lemma~\ref{lemma:A.9}, it follows that
\begin{align*}
\mathbb{E}\left(\left\|\Delta_t\right\|^2 \mid \mathcal{F}_{t-1}\right) & \leqslant\left\|\Delta_{t-1}\right\|^2+4 K \eta_t^2-2 \eta_t K^{-1}\left\|\Delta_{t-1}\right\|^2 \\
& \leqslant\left\|\Delta_{t-1}\right\|^2+4 K \eta_t^2- \eta_t K^{-1}\left\|\Delta_{t-1}\right\|^2 \\
& \leqslant\left(1-\eta_t K^{-1}\right)\left\|\Delta_{t-1}\right\|^2+4 K \eta_t^2 .
\end{align*}
Taking expected value, we conclude that
\begin{align*}
\frac{\mathbb{E}\left(\|\left.\Delta_t\right|^2\right)}{\eta_t} \leqslant \frac{\left(1-\eta_t K^{-1}\right)}{\eta_t} \mathbb{E}\left(\|\Delta_{t-1}\|^2\right)+4 K \eta_t.    
\end{align*}
Finally, using Lemma A.10 in \cite{su2018uncertainty}, we conclude that $\sup _{1 \leqslant t<\infty} \mathbb{E}\left(\left\|\Delta_t\right\|^2\right) / \eta_t<\infty$ and complete the proof.
\end{proof}

\subsubsection{Proof of Lemma~\ref{lemma:A.9}}
\begin{proof}
    Based on $\Delta_t$ defined before, we have
\begin{align*}
\left\|\Delta_t\right\|^2 &\leqslant\left\|\Delta_{t-1}-\eta_t \hat{H}_t\left(\bfQ_{t-1}(S_t,A_t)\right)\right\|^2 \\
& =\left\|\Delta_{t-1}-\eta_t H\left(\bfQ_{t-1}\right)-\eta_t \zeta_t\right\|^2 \\
& =\left\|\Delta_{t-1}\right\|^2+\eta_t^2 \|H\left(\bfQ_{t-1}\right)\|^2+\eta_t^2 \|\zeta_t\|^2 - 2 \eta_t \Delta_{t-1}^{\top} H\left(\bfQ_{t-1}\right) -2 \eta_t \Delta_{t-1}^{\top} \zeta_t \\
& \qquad + 2 \eta_t^2 H^{\top}\left(\bfQ_{t-1}(S_t,A_t)\right) \zeta_t.
\end{align*}

Taking conditional expectation, it follows from Assumption \ref{assumption:A1} that
\begin{align*}
\mathbb{E}\left(\left\|\Delta_t\right\|^2 \mid \mathcal{F}_{t-1}\right) 
& \leqslant\|\Delta_{t-1}\|^2+2 \eta_t^2 \| H\left(\bfQ_{t-1}\right)\|^2+2 \eta_t^2 \mathbb{E}\left(\|\zeta_t\|^2 \mid \mathcal{F}_{t-1}\right) - 2 \eta_t \Delta_{t-1}^{\top} H\left(\bfQ_{t-1}\right) \\
& \stackrel{(\mathrm{i})}{\leqslant} \|\Delta_{t-1}\|^2+2 \eta_t^2 K + 2 \eta_t^2 K-2 \eta_t K^{-1}\| \Delta_{t-1}\|^2 \\
& =\left\|\Delta_{t-1}\right\|^2+4 K \eta_t^2-2 \eta_t K^{-1}\left\|\Delta_{t-1}\right\|^2 \qquad \tag{A.9} \label{eqn:a.9}
\end{align*}
Here (i) is due to Lemmas \ref{lemma:A.7}, \ref{lemma:A.8}, and \ref{lemma:A.1} which is stated below.
\begin{customlemma}{A.3.2}\label{lemma:A.1}
Given $\textbf{Q}^*$ is the unique solution of $H(\mathbf{Q})=0$, then
\begin{align*}
    (\textbf{Q}-\textbf{Q}^*)^\top H(\textbf{Q}) \geq K^{-1}||\textbf{Q}-\textbf{Q}^*||^2
\end{align*}
for all $Q \in \Theta$.
\end{customlemma}
\bigskip

Now, since $\sum_{t=1}^{\infty} \eta_t^2<\infty$, Robbins-Siegmund Theorem implies that $\Delta_t \stackrel{\text { a.s. }}{\longrightarrow} \Delta^*$ for some random vector $\Delta^*$ and
$$
\mathbb{P}\left(\sum_{t=1}^{\infty} 2 \eta_t K^{-1}\left\|\Delta_{t-1}\right\|^2<\infty\right)=1 .
$$

The condition $\sum_{t=1}^{\infty} \eta_t=\infty$ implies that $\Delta_t \stackrel{\text { a.s. }}{\longrightarrow} 0$.
\end{proof}

\subsubsection{Proof of Lemma~\ref{lemma:A.1}}
\begin{proof}
\begin{align*}
\left(\bfQ - \bfQ^* \right)^\top H(\bfQ)
=&(\bfQ-\bfQ^*)^\top\left(\bfQ-\ev(R)-\gamma \ev_{S'}\max_{a'}\bfQ(S', a')\right)\\
=& (\bfQ-\bfQ^*)^\top\left(\bfQ-\ev(R)-\gamma \ev_{S'}\max_{a'}\bfQ^*(S', a') \right.\\ 
 &\qquad \left. -\gamma \ev_{S'}\max_{a'}\bfQ(S', a') +\gamma \ev_{S'}\max_{a'}\bfQ^*(S', a')\right)\\
=&(\bfQ-\bfQ^*)^\top\left(\bfQ-\ev(R)-\gamma \ev_{S'}\max_{a'}\bfQ^*(S', a')\right)\\
    & \qquad -(\bfQ-\bfQ^*)^\top\left(\gamma \ev_{S'}\max_{a'}\bfQ(S', a') -\gamma \ev_{S'}\max_{a'}\bfQ^*(S', a')\right)\\
\geq &\|\bfQ-\bfQ^*\|^2-\|\bfQ-\bfQ^*\| \left\|\gamma \ev_{S'}\max_{a'}\bfQ(S', a') - \gamma \ev_{S'}\max_{a'}\bfQ^*(S', a')\right\|\\
\geq & \|\bfQ-\bfQ^*\|^2-\gamma \|\bfQ-\bfQ^*\| \left\|\bfQ-\bfQ^*\right\|_{\infty}\\
\geq & \|\bfQ-\bfQ^*\|^2-\gamma \|\bfQ-\bfQ^*\|^2\\
\geq & (1-\gamma)\|\bfQ-\bfQ^*\|^2.
\end{align*}
\end{proof}

\subsubsection{Proof of Lemma~\ref{lemma:A.7}}

\begin{proof}
    Since the random reward is uniformly bounded in $[0,1]$ and the Q-value is in a compact set by Assumption~\ref{assumption:A1}, so there is a constant $K$ such that $\|H(\bfQ)(S_t,A_t)\|^2 \leqslant K$ for all $\bfQ \in \Theta$.

    For the second statement, based on the definitions of $\widehat{H}_t(\bfQ_t(S_t,A_t))$ and $H_t(\bfQ)(S_t,A_t)$ we have,
\begin{align*} 
&\ev_\mu\left\{ \left(\widehat{H}_t(\bfQ)(S_t,A_t) -H_t(\bfQ)(S_t,A_t)\right)^T \left(\widehat{H}_t(\bfQ)(S_t,A_t)-H_t(\bfQ)(S_t,A_t)\right)\right\} \\
&= \ev_\mu\left\{||\widehat{H}_t(\bfQ)(S_t,A_t)||^2 + ||H_t(\bfQ)||^2 - \left(\widehat{H}_t(\bfQ)(S_t,A_t)\right)^\top H_t(\bfQ) - (H_t(\bfQ))^\top\left(\widehat{H}_t(\bfQ)(S_t,A_t)\right) \right\}\\
&=\ev_\mu\left\{||(\bfQ(S_t,A_t)-\widehat{\mcT}_{t+1} \bfQ(S_t,A_t))||^2 \right\} - ||H_t(\bfQ)||^2
\intertext{Expanding the norm and simplifying we get,}
&= \ev_\mu\left\{R_{t,1}(S_t,A_t)^\top R_{t,1}(S_t,A_t)\right\} + \frac{B_t - 1}{B_t} \ev_\mu\left\{R_{t,1}(S_t,A_t)\right\}^\top \ev_\mu\left\{R_{t,1}(S_t,A_t)\right\} \\
&\qquad + \frac{\gamma^2}{B_t} \ev_\mu\left\{M_{t,1}(\bfQ)^\top M_{t,1}(\bfQ)\right\} +  \frac{\gamma^2(B_t - 1)}{B_t} \ev_\mu\left\{M_{t,1}(\bfQ)\right\}^\top \ev_\mu\left\{M_{t,1}(\bfQ)\right\} - \bfR^\top \bfR - \gamma^2 M(\bfQ)^\top M(\bfQ) \\
&= B_t^{-1}\left(\ev_\mu\left\{R_{t,1}(S_t,A_t)^\top R_{t,1}(S_t,A_t)\right\} + \gamma^2 \ev_\mu\left\{M_{t,1}(\bfQ)^\top M_{t,1}(\bfQ)\right\} - \ev_\mu\left\{R_{t,1}(S_t,A_t)\right\}^\top \ev_\mu\left\{R_{t,1}(S_t,A_t)\right\} \right.\\
&\qquad \left.- \gamma^2  \ev_\mu\left\{M_{t,1}(\bfQ)\right\}^\top \ev_\mu\left\{M_{t,1}(\bfQ)\right\}\right) \\
&\leq KB_t^{-1}
\end{align*}
where $M_{t,1}(\bfQ)=\max_{a_1'}\bfQ(S'_{t,1}, a_1')$, and the last inequality holds since the rewards and the Q-values are bounded by Assumption \ref{assumption:A1}.

For the last statement, adding $H_t(\bfQ^*)(S_t,A_t)$ to the original terms and simplifying, we get
\begin{align*}
&= \ev_\mu\left\{||\widehat{H}_t(\bfQ)(S_t,A_t) - \widehat{H}_t(\bfQ^*)(S_t,A_t)||^2\right\} - ||H_t(\bfQ)(S_t,A_t) - H_t(\bfQ^*)(S_t,A_t)||^2 \\
&= \ev_\mu\left\{\gamma^2 ||\widehat{M}(\bfQ)(S_t,A_t) - \widehat{M}(\bfQ^*)(S_t,A_t)||^2\right\} - \gamma^2 ||M(\bfQ)(S_t,A_t) - M(\bfQ^*)(S_t,A_t)||^2 \\
&= \frac{\gamma^2}{B_t} \ev_\mu\left\{||M_{t,1}(\bfQ)||^2 + ||M_{t,1}(\bfQ^*)||^2\right\} - \frac{\gamma^2}{B_t}\left(\ev_\mu\left\{M_{t,1}(\bfQ)\right\}^\top \ev_\mu\left\{M_{t,1}(\bfQ)\right\} \right. \\
&\qquad \left. + \ev_\mu\left\{M_{t,1}(\bfQ^*)\right\}^\top \ev_\mu\left\{M_{t,1}(\bfQ^*)\right\} \right) \\
&= B_t^{-1}\ev_\mu\left\{\left\|\bfQ(S_t,A_t) - \mcT_{t+1}(\bfQ)(S_t,A_t) - \bfQ^* + \mcT_{t+1}(\bfQ^*)(S_t,A_t)\right\|^2\right\} \\
&\qquad - B_t^{-1}\left(\left\|\bfQ(S_t,A_t) - \bfQ^*(S_t,A_t)\right\|^2 - \gamma\ev_\mu\left\{\left(\bfQ(S_t,A_t) - \bfQ^*(S_t,A_t)\right)^\top \left(M_{t,1}(\bfQ) - M_{t,1}(\bfQ^*)\right)\right\} \right. \\
& \qquad \qquad \left. -\gamma\ev_\mu\left\{\left(M_{t,1}(\bfQ) - M_{t,1}(\bfQ^*)\right)^\top \left(\bfQ(S_t,A_t) - \bfQ^*(S_t,A_t)\right)\right\} - \gamma^2 \ev_\mu\left\{M_{t,1}(\bfQ)^\top M_{t,1}(\bfQ^*) \right. \right. \\
&\qquad \qquad \qquad \left. \left. + M_{t,1}(\bfQ^*)^\top M_{t,1}(\bfQ)\right\} - \gamma^2 \ev_\mu\left\{M_{t,1}(\bfQ)\right\}^\top \ev_\mu\left\{M_{t,1}(\bfQ)\right\} \right. \\
&\qquad \qquad \qquad \qquad \left. - \gamma^2 \ev_\mu\left\{M_{t,1}(\bfQ^*)\right\}^\top \ev_\mu\left\{M_{t,1}(\bfQ^*)\right\}\right)\\
&\leq KB_t^{-1} h_t(\bfQ)(S_t,A_t).
\end{align*}
where $M_{t,1}(\bfQ)=\max_{a_1'}\bfQ(S'_{t,1}, a_1')$, $M_{t,1}(\bfQ^*)=\max_{a_1'}\bfQ^*(S'_{t,1}, a_1')$, and the last inequality holds since the rewards and the Q-values are bounded by Assumption \ref{assumption:A1}.
\end{proof}

\subsubsection{Proof of Lemma~\ref{lemma:A.8}}
\begin{proof}
    By Lemma \ref{lemma:A.7} and Assumption \ref{assumption:A1}, we have $\mathbb{E}\left(\|\epsilon_t\|^2 | \mathcal{F}_{t-1}\right) \leqslant K B_t^{-1} h_{t-1}\left(\bfQ\right)$, where $K>0$ is some constant. Taking expectation, it follows that $\mathbb{E}_\mu\left(\|\epsilon_t\|^2\right) \leqslant K B_t^{-1} \mathbb{E}_\mu\left\{h_{t-1}\left(\bfQ\right)\right\}$, which is the first statement.
    
The second statement is a direct consequence of Lemma \ref{lemma:A.7}.\\
To prove the third statement, we can utilize Theorem 3 from \cite{rosenthal1970on} which proves the following inequality for $p>2$ and independent, centered Random variables $X_i \in L^p:$
\begin{align*}
E\left[\left|\sum_{i=1}^n X_i\right|^p\right] 
\leq C(p) \max \left\{\sum_{i=1}^n E\left[\left|X_i\right|^p\right],\left(\sum_{i=1}^n E\left[X_i^2\right]\right)^{p / 2}\right\} .
\end{align*}
In our case,
\begin{align*}
\ev_\mu\left(\left\|\zeta_t \right\|^{2k} \right) 
= \ev_\mu\left[\left\|\sum_{i=1}^{B_t} \left(\frac{1}{B_t}\left(R_{t,i}(S_t, A_t)-B_t \bfR\right) + \frac{\gamma}{B_t}\left(\max_{a_i'}Q_t(S'_{t,i}, a_i') -B_t\bfM(\bfQ_t)\right)\right)\right\|^{2k}\right]
\end{align*}
Let us denote the terms inside the summation as $X_i$, then from above result
\begin{align*}
\ev_\mu\left(||\zeta_t||^{2k}\right) 
\leq C(2k) \max\left\{\sum_{i=1}^{B_t} E\left[\left\|X_i\right\|^{2k}\right],\left(\sum_{i=1}^n E\left[X_i^2\right]\right)^k\right\}
&\stackrel{(\mathrm{i})}{\leqslant} C(2k)\left(KB_t^{-1}\right)^{k}\\
&= K_1B_t^{-k}
\end{align*}
where (i) is due to Lemma \ref{lemma:A.7}. This completes the proof.
\end{proof}

\subsubsection{Proof of Lemma~\ref{lemma:A.4}}
\begin{proof}
We first state an auxiliary lemma useful in proving our result.
\begin{customlemma}{A.6.1}\label{lemma:A.3}
Let $a(x)$ be a function on $\mathbb{R}$, and let $k<n$ be two integers.
\begin{enumerate}
    \item If $a(x)$ is non-decreasing, then 
    \begin{align*}
        \int_{k-1}^n a(x) d x \leqslant \sum_{i=k}^n a(i) \leqslant \int_k^{n+1} a(x) d x.
    \end{align*}
    \item If $a(x)$ is non-increasing, then
    \begin{align*}
        \int_k^{n+1} a(x) d x \leqslant \sum_{i=k}^n a(i) \leqslant \int_{k-1}^n a(x) d x.
    \end{align*}
\end{enumerate}
\end{customlemma}
\bigskip

Now, when $\beta=1$, it is easy to prove our result.
W.L.O.G, we may assume $B_t=\ceil{t^{\beta}}$ with $\beta\in (0, 1)$. Since $B_t\geq t^\beta$, it follows from Lemma \ref{lemma:A.3} that
\begin{align*}
\sum_{t=1}^{\ceil{rT}}B_t^{-1}\leq \sum_{t=1}^{\ceil{rT}}t^{-\beta}
=\sum_{t=2}^{\ceil{rT}}t^{-\beta}+1
&\leq \int_1^{\ceil{rT}} t^{-\beta} dt+1\\
&\leq  \int_1^{rT+1} t^{-\beta} dt+1\\
&= \frac{1}{1-\beta}(rT+1)^{1-\beta}-\frac{1}{1-\beta}+1\\
&= \frac{1}{1-\beta}(rT+1)^{1-\beta}+C_1.
\end{align*}

Similarly, since $B_t\leq t^\beta+1$, it holds that
\begin{align*}
\sum_{t=1}^{\ceil{rT}}B_t^{-1}\geq \sum_{t=1}^{\ceil{rT}}\frac{1}{t^\beta+1}
= \sum_{t=1}^{\ceil{rT}}\left\{ \frac{1}{t^{\beta}}-\frac{1}{t^{\beta}}+\frac{1}{t^\beta+1}\right\}
&= \sum_{t=1}^{\ceil{rT}}\left\{ \frac{1}{t^{\beta}}-\frac{1}{t^{\beta}(t^{\beta}+1)}\right\}\\
&=\sum_{t=1}^{\ceil{rT}} t^{-\beta}-\sum_{t=1}^{\ceil{rT}}\frac{1}{t^{\beta}(t^{\beta}+1)}.
\end{align*}
By Lemma \ref{lemma:A.3}, we have
\begin{align*}
\sum_{t=1}^{\ceil{rT}} t^{-\beta}\geq \int_{1}^{\ceil{rT}+1} t^{-\beta}dt
\geq  \int_{1}^{rT} t^{-\beta}dt
=\frac{1}{1-\beta}(rT)^{1-\beta}-\frac{1}{1-\beta}
\end{align*}
and
\begin{align*}
\sum_{t=1}^{\ceil{rT}}\frac{1}{t^{\beta}(t^{\beta}+1)}
\leq \sum_{t=1}^{rT+1}\frac{1}{t^{\beta}(t^{\beta}+1)}
\leq  \sum_{t=1}^{rT+1} \frac{1}{t^{2\beta}}
&=1+\sum_{t=2}^{rT+1} \frac{1}{t^{2\beta}}\\
&\leq 1+\int_1^{rT+1} t^{-2\beta}dt\\
&=1+\frac{1}{1-2\beta}(rT+1)^{1-2\beta}-\frac{1}{1-2\beta}.
\end{align*}
Combining the above inequalities, we show that
\begin{align*}
\sum_{t=1}^{\ceil{rT}}B_t^{-1} &\geq \frac{1}{1-\beta}(rT)^{1-\beta}-\frac{1}{1-2\beta}(rT+1)^{1-2\beta} + C_2.
\end{align*}
Hence, we conclude that
\begin{align*}
\frac{1}{1-\beta}(rT)^{1-\beta} - \frac{1}{1-2\beta}(rT+1)^{1-2\beta}+C_2 
\leq \sum_{t=1}^{\ceil{rT}}B_t^{-1} 
\leq \frac{1}{1-\beta}(rT+1)^{1-\beta}+C_1,
\intertext{and}
\frac{1}{1-\beta}T^{1-\beta}-\frac{1}{1-2\beta}(T+1)^{1-2\beta}+C_2
\leq \sum_{t=1}^{T}B_t^{-1}
\leq \frac{1}{1-\beta}(T+1)^{1-\beta}+C_1.
\end{align*}
Using the above two inequalities, we show that
\begin{align*}
\frac{\sum_{t=1}^{\ceil{rT}}B_t^{-1}}{\sum_{t=1}^{T}B_t^{-1}}
\leq \frac{  \frac{1}{1-\beta}(rT+1)^{1-\beta}+C_1}{\frac{1}{1-\beta}T^{1-\beta}-\frac{1}{1-2\beta}(T+1)^{1-2\beta}+C_2} \to r^{1-\beta} 
\end{align*}
and
\begin{align*}
\frac{\sum_{t=1}^{\ceil{rT}}B_t^{-1}}{\sum_{t=1}^{T}B_t^{-1}}
\geq \frac{\frac{1}{1-\beta}(rT)^{1-\beta}-\frac{1}{1-2\beta}(rT+1)^{1-2\beta}+C_2}{  \frac{1}{1-\beta}(T+1)^{1-\beta}+C_1} \to r^{1-\beta}. 
\end{align*}
By the squeeze theorem, we show that
\begin{align*}
\lim_{T\to \infty}\frac{\sum_{t=1}^{\ceil{rT}}B_t^{-1}}{\sum_{t=1}^{T}B_t^{-1}}=r^{1-\beta}.
\end{align*} 
\end{proof}

\subsubsection{Proof of Lemma~\ref{lemma:A.3}}
\begin{proof}
This is Lemma S.8 in \cite{liu2023statisticalinferencestochasticgradient}.
\end{proof}

\subsubsection{Proof of Lemma~\ref{lemma:A.5}}
\begin{proof}
    This is Lemma S.6 in \cite{liu2023statisticalinferencestochasticgradient}.
\end{proof}

\subsubsection{Proof of Lemma~\ref{lemma:A.10}}

\begin{proof}
    This is a direct consequence of Assumption \ref{assumption:A1} and Lemma \ref{lemma:A.9}.
\end{proof}

\subsection{Additional Results}\label{appendix:d}
In this section, we provide sampling distribution plots for $\beta=0, 0.05, 0.2, 0.3, 0.4$, and 0.5.

\begin{figure}[H]
    \centering
    \includegraphics[width=0.75\linewidth]{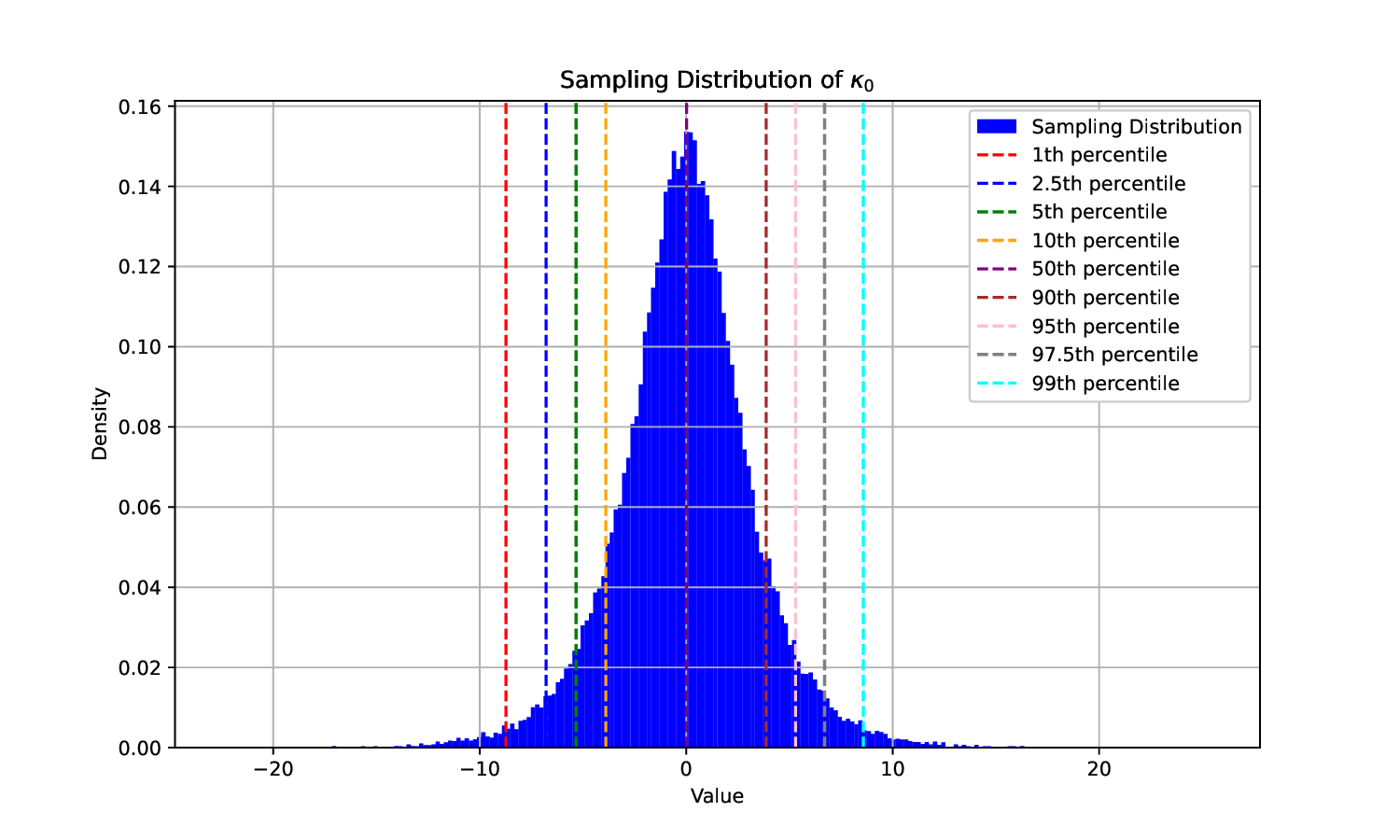}
    \caption{Sampling distribution of $\kappa_\beta$ for $\beta=0$}
    \label{fig:kappa0plot}
\end{figure}


\begin{figure}[H]
    
    \centering
    \includegraphics[width=0.75\linewidth]{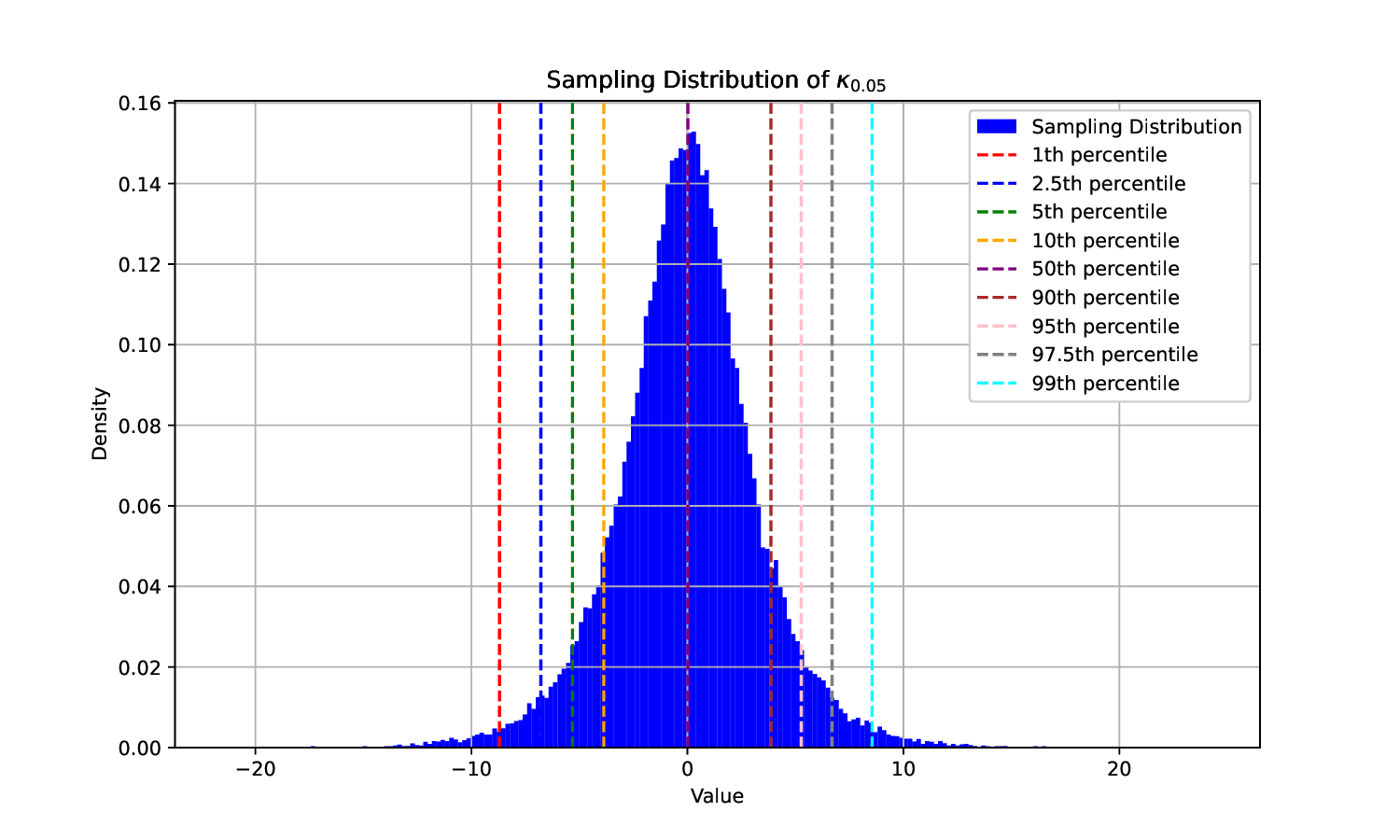}
    \caption{Sampling distribution of $\kappa_\beta$ for $\beta=0.05$}
    \label{fig:kappa0.05plot}
\end{figure}

\begin{figure}[H]
    \centering
    \includegraphics[width=0.75\linewidth]{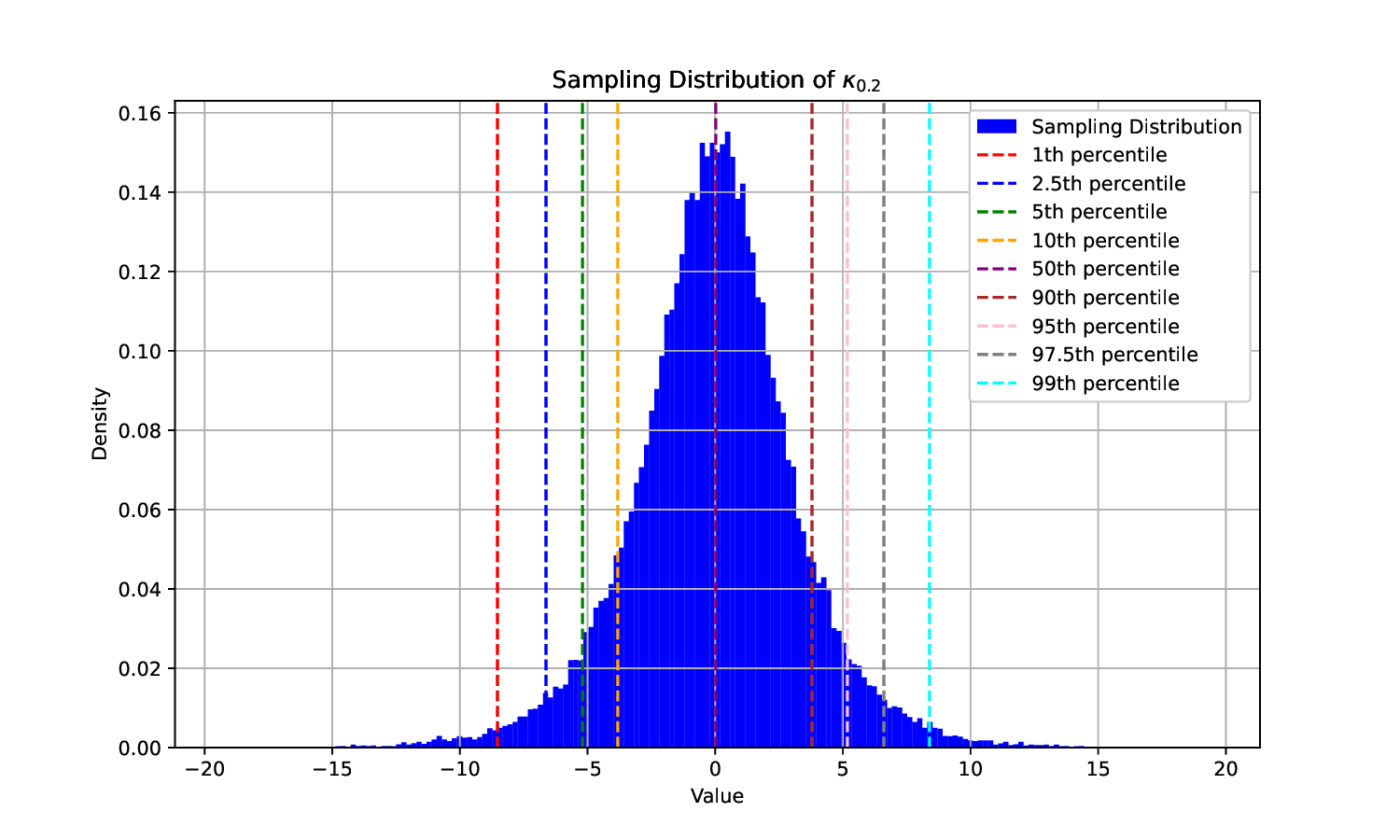}
    \caption{Sampling distribution of $\kappa_\beta$ for $\beta=0.2$}
    \label{fig:kappa0.2plot}
\end{figure}

\begin{figure}[H]
    \centering
    \includegraphics[width=0.75\linewidth]{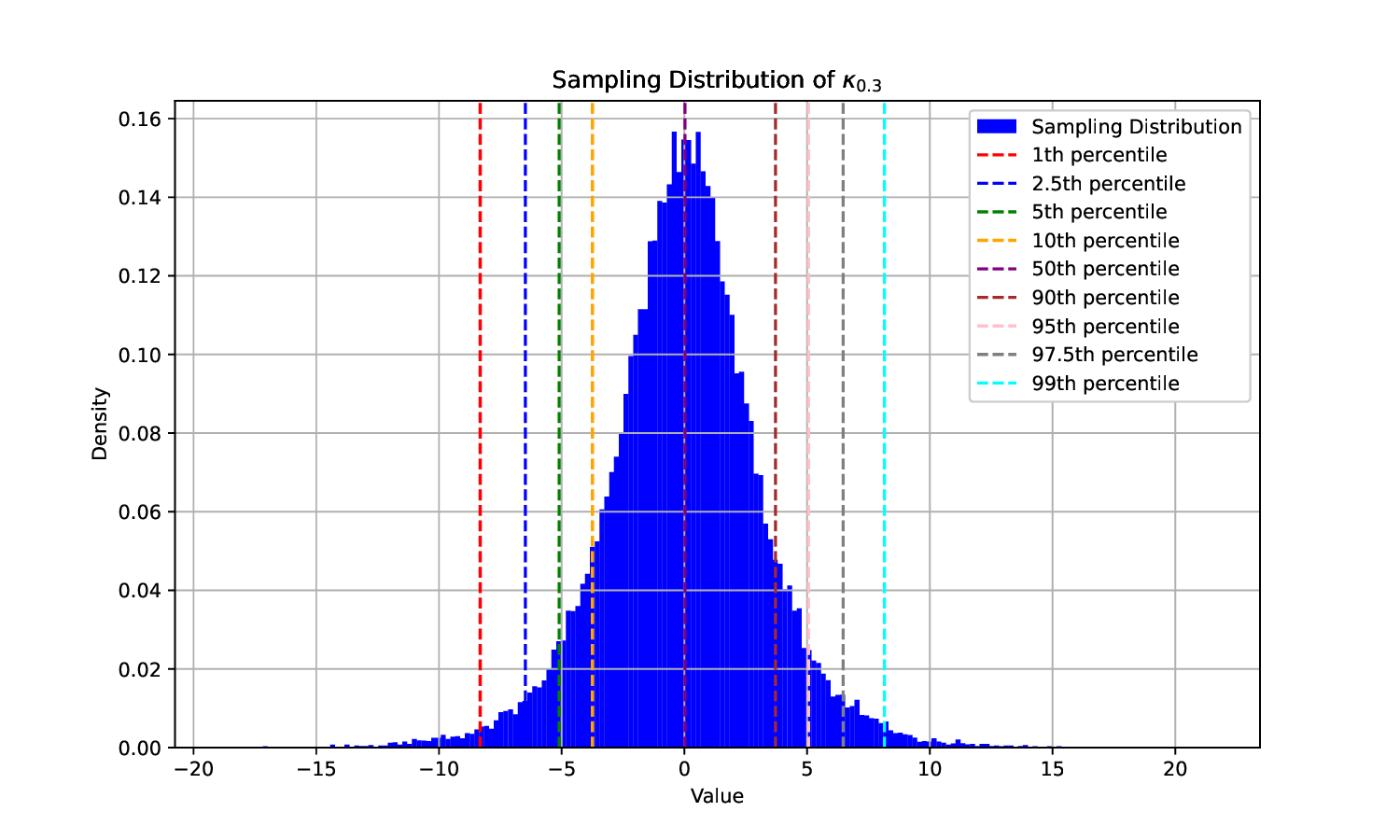}
    \caption{Sampling distribution of $\kappa_\beta$ for $\beta=0.3$}
    \label{fig:kappa0.3plot}
\end{figure}

\begin{figure}[H]
    
    \centering
    \includegraphics[width=0.75\linewidth]{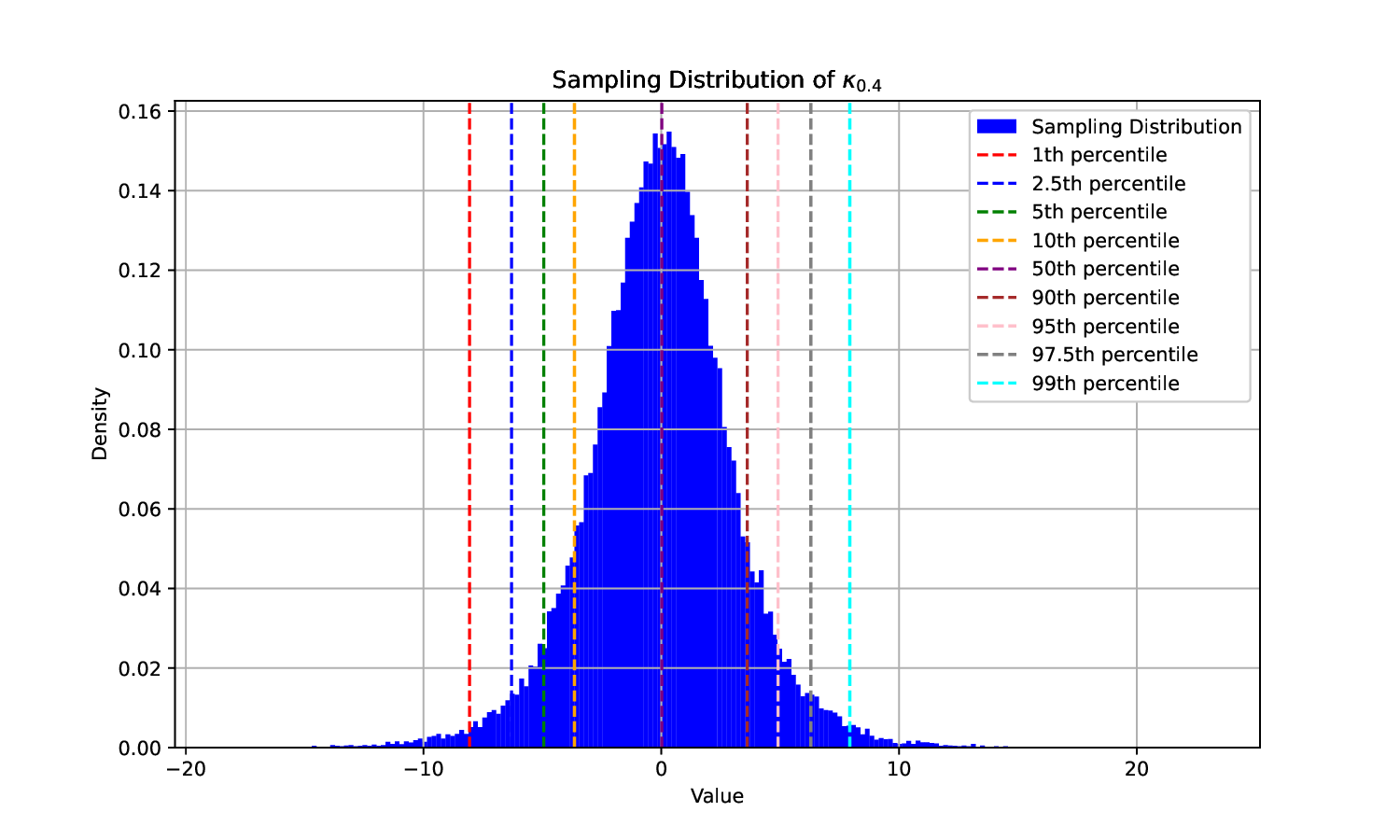}
    \caption{Sampling distribution of $\kappa_\beta$ for $\beta=0.4$}
    \label{fig:kappa0.4plot}
\end{figure}

\begin{figure}[H]
    \centering
    \includegraphics[width=0.75\linewidth]{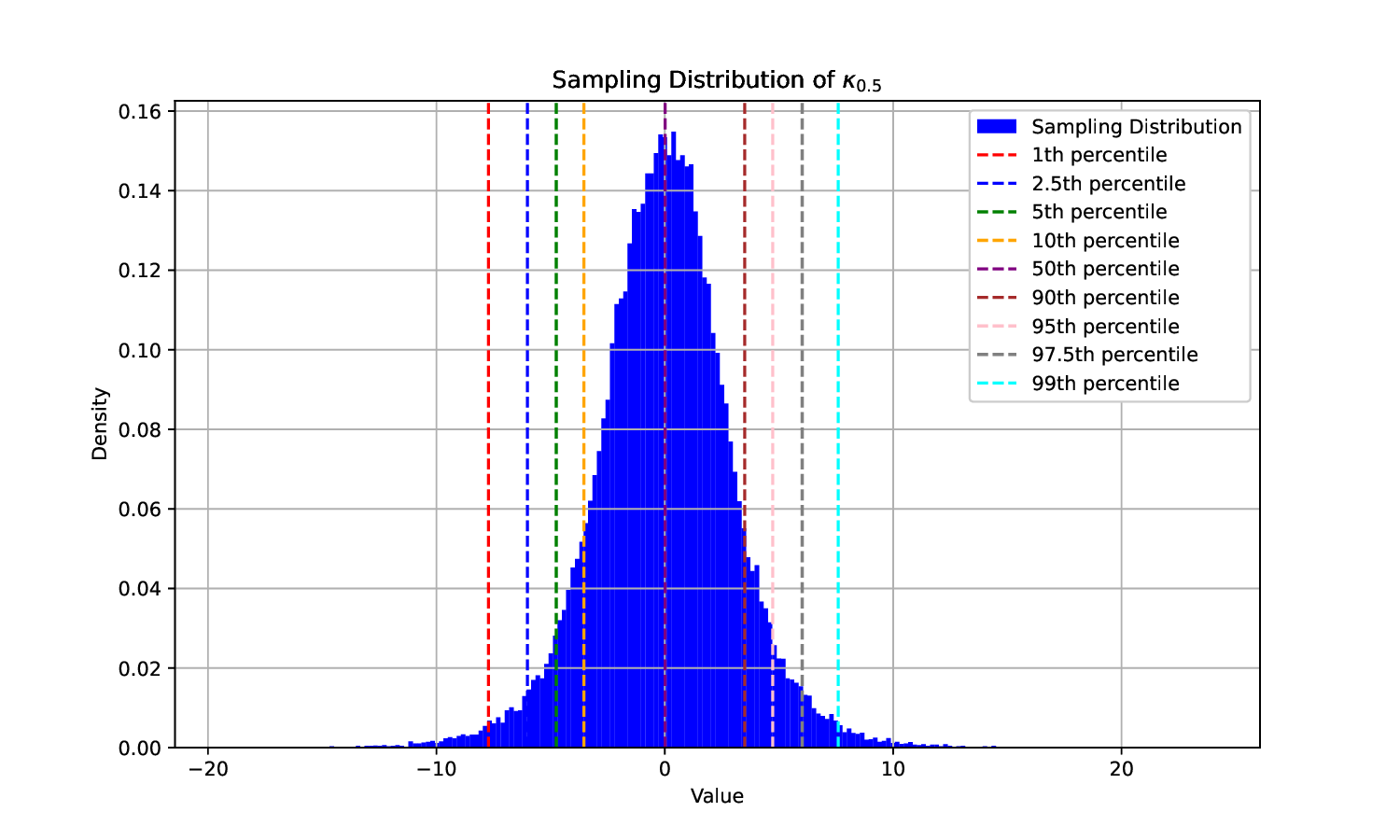}
    \caption{Sampling distribution of $\kappa_\beta$ for $\beta=0.5$}
    \label{fig:kappa0.5plot}
\end{figure}

\bibliographystyle{plain}
\bibliography{Lit}

\end{document}